\documentclass[journal]{IEEEtran}

\usepackage{amssymb}
\usepackage{amsmath,amssymb,amsfonts}
\usepackage{mathtools}
\usepackage{algorithmic}
\usepackage{graphicx}
\usepackage{makecell}
\usepackage{textcomp}
\usepackage{gensymb}
\usepackage{booktabs}
\usepackage{mathrsfs}
\usepackage{color, colortbl} 
\makeatletter
\let\MYcaption\@makecaption
\makeatother

\usepackage[font=footnotesize]{subcaption}

\makeatletter
\let\@makecaption\MYcaption
\makeatother
\usepackage{comment}

\begin{document}
%
\title{Efficient Unsupervised Domain Adaptation Regression for Spatial-Temporal Sensor Fusion}
%
%
%

\author{Keivan~Faghih Niresi,
        Ismail~Nejjar,
        and~Olga~Fink,~\IEEEmembership{Member,~IEEE}
\thanks{Manuscript received ...}
\thanks{Corresponding author: Olga Fink (olga.fink@epfl.ch)}
\thanks{Keivan Faghih Niresi, Ismail Nejjar, and Olga Fink are with the Intelligent Maintenance
and Operations Systems Laboratory, EPFL, 1015 Lausanne, Switzerland
(e-mail: keivan.faghihniresi@epfl.ch; ismail.nejjar@epfl.ch; olga.fink@epfl.ch).}
\thanks{Copyright (c) 2025 IEEE. Personal use of this material is permitted. However, permission to use this material for any other purposes must be obtained from the IEEE by sending a request to pubs-permissions@ieee.org}
}

%
%

\markboth{IEEE Internet of Things Journal}%
{Shell \MakeLowercase{\textit{et al.}}: Bare Demo of IEEEtran.cls for IEEE Journals}
%



\maketitle

\begin{abstract}
The growing deployment of low-cost, distributed sensor networks in environmental and biomedical domains has enabled continuous, large-scale health monitoring. However, these  systems often face challenges  related to degraded data quality caused by  sensor drift, noise, and insufficient  calibration -- factors that limit their reliability in real-world applications. Traditional machine learning methods for sensor fusion and calibration rely on extensive feature engineering and struggle to capture spatial-temporal dependencies or adapt to distribution shifts across varying  deployment conditions. To address these challenges, we propose a novel  unsupervised domain adaptation (UDA) method tailored for regression tasks. Our proposed method integrates effectively  with Spatial-Temporal Graph Neural Networks and leverages the alignment of perturbed inverse Gram matrices between  source and target domains, drawing  inspiration from Tikhonov regularization. This approach enables scalable and efficient domain adaptation without requiring labeled data in the target domain. We validate our novel method on real-world datasets from two distinct applications: air quality monitoring and EEG signal reconstruction. Our method achieves state-of-the-art performance which paves the way for more robust and transferable sensor fusion models in both environmental and physiological contexts. Our code is
available at https://github.com/EPFL-IMOS/TikUDA.
\end{abstract}


%
\IEEEpeerreviewmaketitle

\section{Introduction}
\label{sec:introduction}

The growing deployment  of low-cost, distributed sensor networks  has fundamentally reshaped how real-time data is collected and utilized across a wide range of applications -- from urban air quality monitoring to physiological sensing. In environmental contexts, sensor arrays are deployed to measure key pollutants such as carbon monoxide (CO), nitrogen dioxide (NO\textsubscript{2}), and tropospheric ozone (O\textsubscript{3}), to enabling data-informed public health interventions, exposure alerts, and evidence-based urban planning \cite{lu2019meteorology,bibri2024smarter}. In biomedical domains, wearable sensors like electroencephalography (EEG) continuously record physiological signals to support clinical diagnostics, brain–computer interfaces, and patient monitoring. Together, these systems underpin efforts to protect both public health and the environment by enabling timely, data-driven decision-making across very different but equally critical domains.

The adoption of Internet of Things (IoT) architectures has enabled the large-scale deployment of low-cost sensor networks for real-time monitoring in both environmental and biomedical domains \cite{lau2019survey,bibri2023environmentally}. However, these low-cost sensing platforms suffer from several interrelated accuracy issues. In air quality networks, gas sensors are rarely calibrated under true field conditions, resulting in biases and drift as ambient temperature, humidity, and pollutant concentrations fluctuate \cite{lewis2018low,lau2017sensor,pau1}. In physiological monitoring, wearable electrodes and inexpensive biosensors exhibit signal degradation due to electrode displacement, motion artifacts, and pronounced inter-subject variability, which together induce degradation of the recorded data. To mitigate these effects, environmental systems often incorporate auxiliary temperature and humidity sensors, while biomedical setups may co-register signals with higher-precision reference instruments or additional modalities \cite{cross2017use}. Although these measures can improve raw measurement fidelity, they introduce significant hardware complexity, increase maintenance demands, and do not fully eliminate cross-sensitivity or calibration inconsistencies. Consequently, both regulatory applications (e.g., pollutant compliance monitoring) and clinical use cases (e.g., patient diagnostics) remain constrained by limited sensitivity and variable data quality \cite{giordano2021low}. Two principal strategies have emerged to address these limitations: hardware redundancy—deploying overlapping low-cost sensors to average out individual device errors—and advanced multisensor fusion algorithms that jointly process multiple data streams to produce robust, high-precision estimates without exclusive reliance on expensive reference equipment \cite{ferrer2024data}.

Traditional machine learning algorithms such as support vector regression (SVR), random forest (RF), and k-nearest neighbors (KNN) have been shown to be feasible options for multisensor fusion and calibration in IoT-based monitoring platforms \cite{pau1}. Despite their fairly good performance, these traditional machine learning models still face challenges in achieving reliable estimation of air pollutants,
limiting their deployment reliability. One limitation of these models is their dependency on extensive feature engineering before data input. For instance, sensor networks often experience multicollinearity, where multiple sensors provide highly correlated information. This redundancy not only diminishes  model performance but also leads to inefficient training processes. Effective feature engineering in such cases requires substantial domain knowledge and can be highly specialized, presenting challenges in developing reliable machine learning models. Moreover, these models typically fail to account for the spatial and temporal interactions among sensor measurements that are particularly relevant in environmental or biomedical sensor networks. Although temporal pattern-based denoising can improve performance as a post-processing step following fusion and calibration, as indicated in \cite{allka}, there remains a critical need for a model that can jointly capture both spatial and temporal features in a comprehensive, end-to-end approach.

Spatial-temporal graph neural networks (STGNNs) have emerged as powerful tools in various applications due to their ability to capture complex spatial-temporal relationships within data \cite{jin2023spatio}. This versatility has been demonstrated in different fields such as traffic data analysis \cite{Cui2020graph, jiang2022graph, bui2022spatial}, industrial IoT \cite{niresi2024physicsenhanced, zhao2024dyedgegat}, and evolving social networks \cite{min2021stgsn}. Leveraging their capability to capture spatial-temporal interactions among sensor measurements, a recent study applied STGNNs for multisensor fusion and calibration in IoT air pollution monitoring platforms \cite{niresi2023spatial}. This study showed that this architecture can achieve promising results compared to other data-driven methods that did not leverage the graph structure of air quality sensor networks. Similarly, STGNNs have been shown to effectively model the spatial relationships among EEG channels and the temporal dynamics of neural activity, leading to improved classification performance in EEG‐based tasks, as highlighted in a recent survey of STGNN applications for EEG signal processing \cite{klepl2024graph}.
STGNNs typically perform well when the testing data distribution aligns with the training data distribution. However, in real-world scenarios, especially in diverse urban environments and different subjects, data distributions are prone to change over time. This shift can result in performance degradation. This occurs because most current data-driven models assume that the statistical properties of data remain the same \cite {ben2006analysis}. In air quality monitoring platforms, geographical variations due to factors such as weather conditions, pollution sources, and urban planning can cause significant differences in data distributions. Similarly, in EEG applications, distribution shifts arise from inter‐subject variability, differences in electrode placement, individual neurophysiology, and hardware configurations, leading to degraded model performance when transferring across users or recording setups. Consequently, models trained in one location or subject may not perform when applied in another. 
A straightforward but expensive solution would be to collect new labeled data at each specific subject or measurement location and train a separate model for each setting. However, this approach depends on deploying high-precision reference equipment such as regulatory-grade gas analyzers for each air quality station or high density EEG systems for every individual to achieve accurate calibration, which is prohibitively expensive. Consequently, there is a critical need for methods that can leverage labeled data from a source domain and adapt models to new target domains whether across different geographic locations or across subjects without requiring costly reference instruments at each deployment.

To address this challenge, unsupervised domain adaptation (UDA) shows promise in transferring knowledge from a labeled source domain to a related but unlabeled target domain \cite{wilson2020survey}. UDA would allow for the utilization of only a single high-precision sensor in the source domain, eliminating the need for labeled data in the target domain. As a result, the need to deploy and manage multiple high-precision sensors across different environments or subjects is eliminated, reducing overall costs.

Various UDA approaches have been applied to time series data to minimize discrepancies across domains by using domain-invariant temporal features extracted through convolutional neural networks (CNNs) and long short-term memory (LSTM) \cite{cai2021time, liu2021adversarial}. However, sensor networks, exhibit spatial-temporal interdependencies among sensors that must be considered for effective fusion and adaptability. Therefore, when applying UDA to spatial-temporal data, it is crucial to address both temporal and spatial interactions. For graph-structured data, minimizing the maximum mean discrepancy (MMD) loss \cite{long2015learning} has been shown to generate transferable node embeddings suitable for cross-network learning tasks \cite{shen2020adversarial}. Additionally, domain adversarial learning strategies \cite{ma2019gcan, zhu2021cross, ganin2016domain} combined with graph convolutional networks (GCNs) as feature extractors have demonstrated improved adaptability across different domains by generating domain-invariant node representations. However, UDA methods for graph-structured data are predominantly designed for classification tasks and often overlook the temporal dependencies of sensor (node) values. This limitation reduces their effectiveness in multisensor fusion and calibration tasks, which require models to predict continuous values \cite{chen2021representation, nejjar2023dare}.

Recently, novel UDA methods have emerged with a focus on regression tasks. Notably, Representation Subspace Distance for Domain Adaptation Regression (RSD) \cite{chen2021representation} aims to align orthogonal bases while preserving the feature scale. The RSD-based approach has shown improved results, highlighting the need for UDA methods specifically designed for regression tasks.
Alternatively, the DARE-GRAM \cite{nejjar2023dare} method approaches UDA for regression from a regressor's perspective. Motivated by the closed-form least square solution, this method proposes aligning the pseudo-inverse of the Gram matrix instead of directly aligning the features. DARE-GRAM, which focuses on scale and angle alignment based on the Gram matrix, results in better-calibrated features for the regressor and exhibits less sensitivity to batch size compared to RSD. Although these two subspace-based approaches \cite{chen2021representation, nejjar2023dare} provide state-of-the-art results for UDA in regression tasks, each training iteration requires computationally intensive singular value decomposition (SVD), which can be computationally too expensive for large-scale sensor networks.

To address existing limitations, this study proposes a novel UDA method for regression tasks that delivers state-of-the-art results while being faster and more scalable. 
Specifically, we introduce a Tikhonov-regularized least squares-based unsupervised deep domain adaptation (TikUDA) method. This approach aligns the inverse of the Gram matrix of the features, perturbed by a scaled identity matrix known as the Tikhonov matrix. This perturbation reduces the condition number of the Gram matrix. The full-rank properties of the Tikhonov matrix allow us to use faster decomposition methods rather than relying on singular value decomposition, ensuring stable training and reducing training time. This is particularly relevant for graph neural networks (GNNs) as the number of nodes (sensors) increases. As the network expands, more sensors are aggregated in the embedding space, thus increasing its dimensionality.

Building on this foundation, the primary goal of this paper is to develop a robust methodology for multisensor fusion and regression across both environmental and physiological domains. Our approach leverages the proposed TikUDA method to bridge the gap between two domains: training on labeled data from low-cost sensors calibrated with a high-precision sensor in one location or subject (the source domain) and transferring that knowledge to a different location or subject (target domain) using unlabeled data from low-cost sensors.

To capture both spatial and temporal dependencies in sensor measurements, we propose using spatial-temporal graph neural networks (STGNNs) within sensor networks. The model first encodes input data into a hidden space with a linear encoder. Temporal dependencies are then extracted using a Gated Recurrent Unit (GRU), followed by spatial pattern extraction using a Graph Attention Network (GAT). To ensure the model generalizes well to the target location, we apply our TikUDA method to align the feature representations from both the source and target domains.

Using real-world datasets, we evaluate our framework on two different tasks. First, in air pollutant estimation, our method accurately predicts concentrations of pollutant at unseen locations using only low-cost sensor data, thereby reducing reliance on high-precision reference instruments and lowering both deployment cost and system complexity. Second, in EEG signal processing, the approach successfully reconstructs a missing channel from neighboring electrodes, enhancing signal fidelity without additional hardware.

The key contributions of this article are as follows:

 \begin{enumerate}
   \item We propose a novel UDA method tailored for spatial-temporal regression tasks that provides state-of-the-art results with improved scalability and faster execution.
  \item We proposed a new UDA method for regression tasks by utilizing the well-known closed-form solutions from Tikhonov regularization, commonly used in linear inverse problems.
   \item We integrate this UDA method with STGNNs, effectively capturing both temporal dynamics and graph-structured data, demonstrating the strength of this combined approach.
   \item We evaluate the proposed method on two real-world sensor networks, demonstrating its practical applicability and superior performance in real-world applications.
 \end{enumerate}

The remainder of the paper is organized as follows: Section \ref{sec:problem_def} provides an in-depth introduction to the relevant background in multisensor fusion and unsupervised domain adaptation for regression tasks. In Section \ref{sec:methodology}, we present the proposed spatial-temporal feature extractor along with our proposed TikUDA method for unsupervised domain adaptation. Sections \ref{sec:setup} and \ref{sec:result} detail the experimental setup and results, illustrating the efficacy of our approach compared to various baseline and state-of-the-art methods. Finally, Section \ref{sec:conclusion} summarizes the findings and suggests possible directions for future research.

\section{Problem Definition}
\label{sec:problem_def}

\subsection{Multisensor Fusion and Calibration}
\begin{figure*}[t]
    \centering
    \begin{subfigure}[t]{\textwidth}
        \centering
        \includegraphics[width=\linewidth]{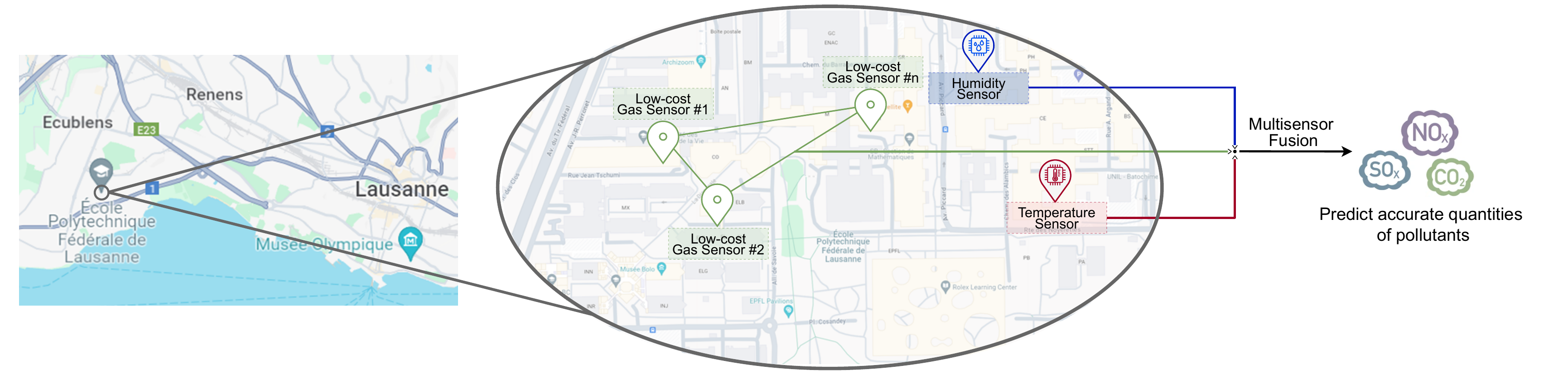}
        \caption{Multisensor fusion in IoT air quality monitoring to predict accurate quantities of various pollutants}
        \label{fig:airquality_fusion}
    \end{subfigure}
    \hfill
    \begin{subfigure}[t]{0.7\textwidth}
        \centering
        \includegraphics[width=\linewidth]{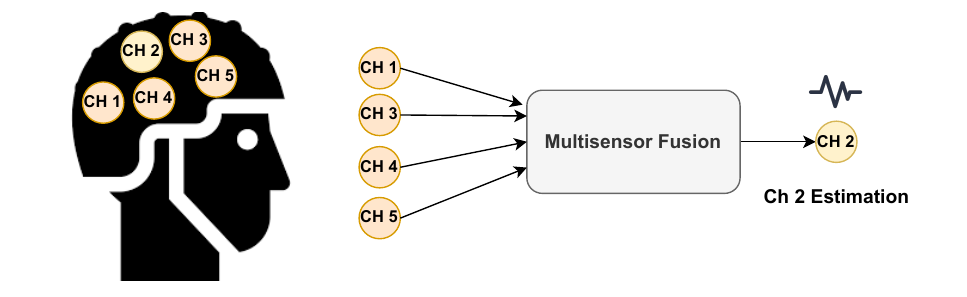}
        \caption{Sensor fusion for estimating missing EEG channel (Channel 2).}
        \label{fig:eeg_fusion}
    \end{subfigure}
    \caption{Overview of sensor fusion tasks across different applications. (a) Air quality (b) EEG signal processing}
    \label{fig:fusion_overview}
\end{figure*}

Multisensor data fusion and calibration involve the integration of data from multiple sensors, denoted as $\{s_1, ..., s_n\}$. In air quality monitoring, these sensors record pollutants, temperature, and humidity; in EEG applications, electrodes capture neural signals across multiple scalp locations. The aim is to produce reliable estimates of target variables by aggregating multiple sensors, as illustrated in Figure \ref{fig:fusion_overview}. 

Given the spatial‐temporal nature of the sensor networks, recent work has applied STGNNs to model sensor interactions and temporal evolution \cite{jin2023spatiotemporal}. STGNNs leverage the graph structure—connecting co‐located gas sensors or neighboring EEG electrodes—to jointly learn spatial dependencies and temporal patterns, outperforming methods that treat these aspects separately.

However, relying solely on direct predictions from low-cost IoT air-quality sensors or on simple interpolation of EEG signals is inadequate for accurate multisensor fusion when models are deployed in new environments or on new subjects. In our approach, we leverage labeled source-domain data from low-cost sensor networks co-located with high-precision reference monitors for air-quality monitoring and from complete EEG recordings collected in an initial subject cohort, while treating as the target domain the unlabeled low-cost sensor measurements obtained in different geographic locations and the incomplete EEG recordings with a missing channel in new participants. Although both source and target domains maintain the same number of sensors or electrodes, distributional shifts arise from local environmental factors or inter-subject variability. To address this, we proposed new domain adaptation methods to align spatial–temporal feature representations across domains. This framework enables accurate estimation of pollutant concentrations and robust reconstruction of missing EEG channels without requiring high-precision instruments or full electrode setups at every deployment.

\subsection{Unsupervised Domain Adaptation}

Unsupervised Domain Adaptation (UDA) aims to overcome the distribution shift between a labeled source domain and an unlabelled target domain. Especially, we aim to learn a model $f_\mathbf{\theta}$ that generalizes well to the target domain. In other words, we aim to minimize the expected error on the target data:
\begin{equation}
\underset{\mathbf{\theta}}{\arg\min} \, \mathbb{E}\{\|f_{\mathbf{\theta}}(\mathbf{x}^{tgt}) - y^{tgt}\|_2^2\},
\end{equation}
where ${y}^{tgt}$ denotes the ground truth label for the target input feature $\mathbf{x}^{tgt}$ and is not available during training.

Different methods aim to minimize the learned feature discrepancy between the source and target domain. Formally, the model \( f_{\mathbf{\theta}}(.) \), can be represented by a feature encoder \( h_{\theta}(.) \) and a regressor \( g_{\theta}(.) \) that maps the learned feature representation for the input data \( \mathbf{z} = h_{\theta}(\mathbf{x}) \) to the predicted pollutant level \(\hat{y} = g_{\theta}(\mathbf{z}) \). During training, the feature matrix for a batch size \( b \) can be expressed as \( \mathbf{Z} = [\mathbf{z_1}, \ldots, \mathbf{z_b}] \), where \( \mathbf{Z} \in \mathbb{R}^{b \times p} \) with \( p \) the feature dimension.

DARE-GRAM \cite{nejjar2023dare}, motivates aligning the inverse Gram matrix (\(\mathbf{Z}^\mathsf{T}\mathbf{Z})^{-1}\) from the well-known closed-form solution of the ordinary least-squares (OLS) to ensure a feature representation that is similar from a regressor perspective. However, in deep learning models, the batch size is often smaller than the feature dimension, i.e., \(b < p\), resulting in a feature matrix \( \mathbf{Z} \in \mathbb{R}^{b \times p} \). Consequently, the Gram matrix (\(\mathbf{Z}^\mathsf{T}\mathbf{Z} \in \mathbb{R}^{p \times p}\)) may have a rank \( r \leq b \), leading to an ill-conditioned matrix that is not invertible. To address this, DARE-GRAM proposed using the Moore-Penrose pseudo-inverse, which generalizes the concept of matrix inversion for matrices that may not be invertible. Additionally, a threshold is applied to singular values to focus on a selected subspace of the inverse Gram matrix. However, the practical implementation of such an algorithm relies on two singular value decompositions, which is computationally expensive. In this paper, we propose an improved method that is less computationally expensive.

\section{Methodology}
\label{sec:methodology}

In this work, we propose a Tikhonov-regularized least squares-based unsupervised domain adaptation (TikUDA) method. This method improves over previous methods, as detailed in Section \ref{sec:TikUDA}. Furthermore, we introduce an STGNN designed to effectively capture both spatial and temporal interactions in multisensor fusion tasks. This model leverages the spatial correlations between different sensor nodes and temporal dependencies within their measurements, providing a promising framework for integrating and analyzing information from multiple sensors over time, as detailed in Section \ref{sec:STGNN}. In Section \ref{sec:overall}, we provide a comprehensive overview and summary of the overall framework.

\subsection{TikUDA}
\label{sec:TikUDA}

\begin{figure}
\centerline{\includegraphics[width=\linewidth]{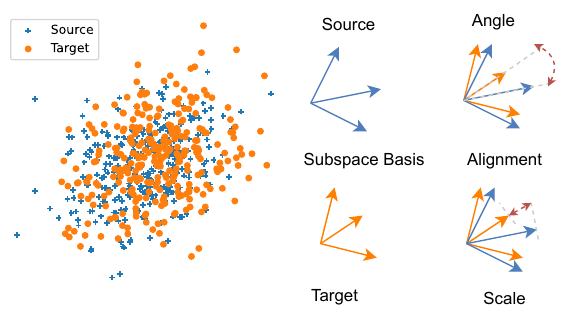}}
\caption{Illustration of subspace-based domain adaptation. The source and target feature distributions differ in both scale and angle. The goal is to align the subspaces.}
\label{motiv}
\end{figure}

To alleviate the ill-conditioned problem of the Gram matrix, we propose using the closed-form solution of Ridge regression -- a specific case of Tikhonov-regularized least-squares -- instead of using the closed-form solution of OLS. Specifically, we apply the Tikhonov matrix 
\( \mathbf{Z}^\mathsf{T} \mathbf{Z} + \alpha \mathbf{I} \), which ensures the matrix is fully ranked by adding the identity matrix. This addition guarantees the smallest eigenvalue is never zero but it depends on \( \alpha \). This eliminates the need to compute the SVD to threshold degenerate spaces (small eigenvalues) and allows for direct inverse matrix decomposition. Following DARE-GRAM methodology \cite{nejjar2023dare}, we propose aligning the subspaces of the source and target domains using the angle between the columns of the inverse Tikhonov matrix, which are ordered based on the importance of their eigenvalue. To ensure consistent scaling between the source and target feature representations, we align their eigenvalues, addressing the scale sensitivity in regression models as highlighted in \cite{chen2021representation}. Figure \ref{motiv} illustrates the main motivation behind TikUDA. It presents two distributions representing features from the source and target domains. These feature representations differ in both scale and orientation, highlighting the challenges of aligning domains with different properties. The primary goal of this work is to align these distributions by adjusting both the scale and the angle of the subspaces they span. The alignment of Gram matrices provides a principled approach to mitigating domain shift by leveraging stable relational structures in representation space. The Gram matrix, capturing pairwise inner products, implicitly encodes the covariance-like structure of the data. While raw features are sensitive to domain-specific perturbations (e.g., sensor noise or background variations), these high-level relational geometries, including inter-vector distances and angles, tend to remain more invariant across domains. Mid-layer Gram matrices in deep networks further exhibit beneficial low-rank structure, concentrating meaningful variance into a few dominant directions while naturally filtering out noise and domain artifacts \cite{skean2025layer}. Aligning Gram matrices thus performs implicit manifold alignment, matching the intrinsic geometry of source and target feature spaces rather than exact feature values. This alignment enables more effective knowledge transfer between domains, which is crucial for improving performance in domain adaptation tasks.

\noindent\textbf{Angle Alignment:}
The inverse Tikhonov matrices for the source and target domains are represented as:
\begin{align}
  \mathbf{G}_s^{-1} &= (\mathbf{Z}_{\text{src}}^\mathsf{T} \mathbf{Z}_{\text{src}} + \alpha \mathbf{I})^{-1} \\
  \mathbf{G}_t^{-1} & = (\mathbf{Z}_{\text{tgt}}^\mathsf{T} \mathbf{Z}_{\text{tgt}} + \alpha \mathbf{I})^{-1} 
\end{align}
respectively. Since the Tikhonov matrix is symmetric and positive definite, its inverse can be computed more efficiently using methods such as Cholesky decomposition \cite{boyd2004convex}. This approach reduces the computational complexity of our proposed algorithm, making it more practical for larger matrices.

Previous methods utilized cosine similarity to measure the angle difference between the source and target using the column space of \( \mathbf{G}_s^{-1} \) and \( \mathbf{G}_t^{-1} \), and is defined as follows:

\begin{equation}
\cos(\phi^{S \leftrightarrow T}_{(i)}) = \frac{{\mathbf{G}_{s,i}^{-1}} \cdot \mathbf{G}_{t,i}^{-1}}{\|\mathbf{G}_{s,i}^{-1}\| \cdot \|\mathbf{G}_{t,i}^{-1}\|},
\label{eq:cosine}
\end{equation}
where \( i \in [1, p] \) represents the \( i \)-th column (out of a total of \( p \) columns) of the inverse Tikhonov matrix. A major drawback of using cosine similarity is that when the angle between two vectors is small, the cosine value approaches 1, leading to a loss of precision and sensitivity in distinguishing between vectors that are nearly parallel. This can pose challenges in applications where subtle differences are crucial, making it difficult to effectively compare them. To address this issue, we introduce the haversine similarity (HS), defined as follows:
\begin{equation}
\mathsf{HS}(\phi^{S \leftrightarrow T}_{(i)}) = 1 - \sqrt{\frac{1 - \cos(\phi^{S \leftrightarrow T}_{(i)})}{2}}.
\label{eq:haversine}
\end{equation}

\begin{figure}
\centerline{\includegraphics[width=0.8\linewidth]{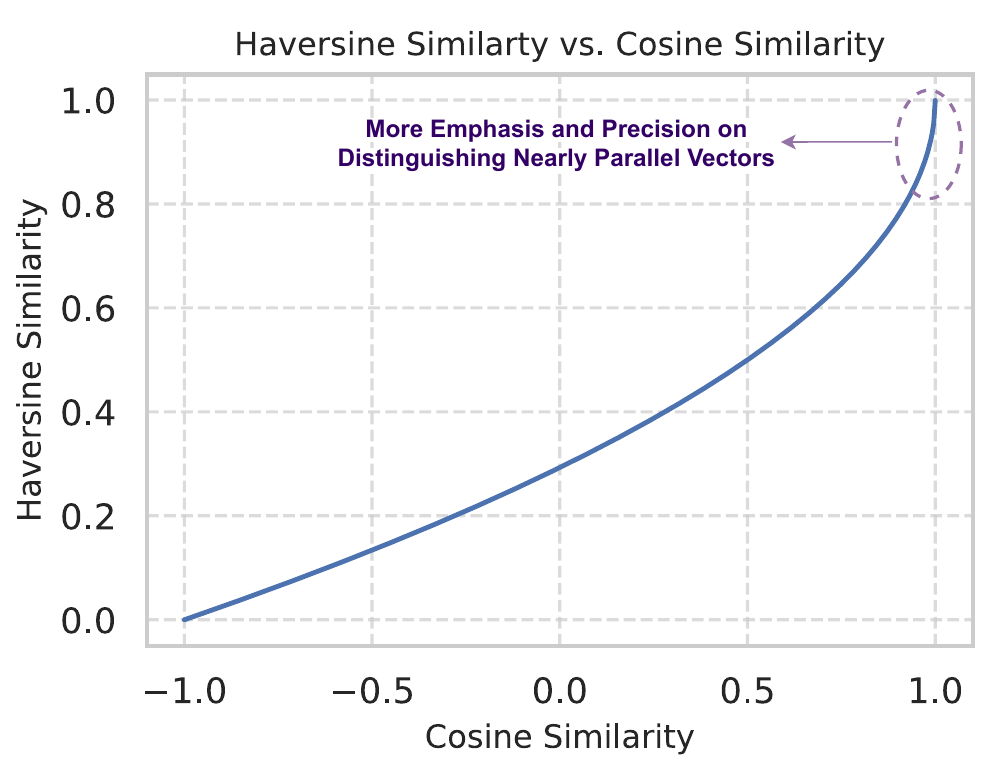}}
\caption{Comparison of cosine similarity and the haversine similarity function.}
\label{similaritycomparison}
\end{figure}

The proposed haversine similarity metric effectively penalizes differences in angles, especially for small ones. We argue that the haversine similarity is more suitable for subspace alignment, as it enforces closer angular alignments between the columns of the inverse Tikhonov matrices of the source and target domains. Figure \ref{similaritycomparison} illustrates the differences between these two metrics for clearer understanding.

The HS values between the subspace spans of the source and target features are assigned to \( \mathbf{m} = [\mathsf{HS}(\phi^{S \leftrightarrow T}_{(1)}), \ldots, \mathsf{HS}(\phi^{S \leftrightarrow T}_{(p)})] \). Thus, the loss function to align the bases from the inverse Tikhonov matrix can be expressed as:
\begin{equation}
L_{\text{angle}} = \| \mathbf{1} - \mathbf{m}\|_1
\label{angle}
\end{equation}

where \( \mathbf{1} \) is a vector of ones with shape \( p \). Minimizing this loss function (\ref{angle}) maximizes the haversine similarity between the representation subspaces of the source and target by reducing the angle between the bases of both domains.

\noindent\textbf{Scale Alignment :}
In addition to angle alignment, which was discussed in the previous subsection, maintaining feature scale consistency across the source and target domains is essential for effective UDA in regression tasks. We find that in regression scenarios, the largest principal eigenvalue ($\lambda_{\text{max}}$) has the most significant impact on feature scaling. Hence, in contrast to the DARE-GRAM \cite{nejjar2023dare}, which selects k-principal eigenvalues for scaling through singular value decomposition (SVD), we only consider the largest eigenvalue of the feature matrix. Our approach avoids the computationally intensive SVD procedure typically used to identify the k-principal eigenvalues. Instead, we can efficiently compute the largest eigenvalue using methods such as the Lanczos algorithm and power methods \cite{lanczos1950iteration}. Thus, the regularization of the scale alignment between the features of the source and target is achieved by minimizing the discrepancy between their first principal eigenvalue as:
\begin{equation}
L_{\text{scale}} = (\lambda_{\text{max}}^{src} - \lambda_{\text{max}}^{tgt})^2.
\label{scale}
\end{equation}

\subsection{Spatial-Temporal Feature Extractor}
\label{sec:STGNN}
To effectively capture both spatial and temporal features, current STGNNs utilize two types of architectures: time-then-space (factorized) and time-and-space (coupled) \cite{jin2023survey}. In time-then-space designs, temporal processing occurs before spatial processing. Both theoretical and empirical evidence suggest that time-then-space architectures, when using GNNs components constrained to the 1-Weisfeiler-Lehman (1-WL) power \cite{xu2018how}, provide a performance advantage compared to time-and-space architectures that utilize the same 1-WL GNN components \cite{gao2022equivalence}. Given these advantages, we opt for the time-then-space architecture for multisensor fusion to achieve more accurate predictions.

\begin{figure*}
\centerline{\includegraphics[width=\linewidth]{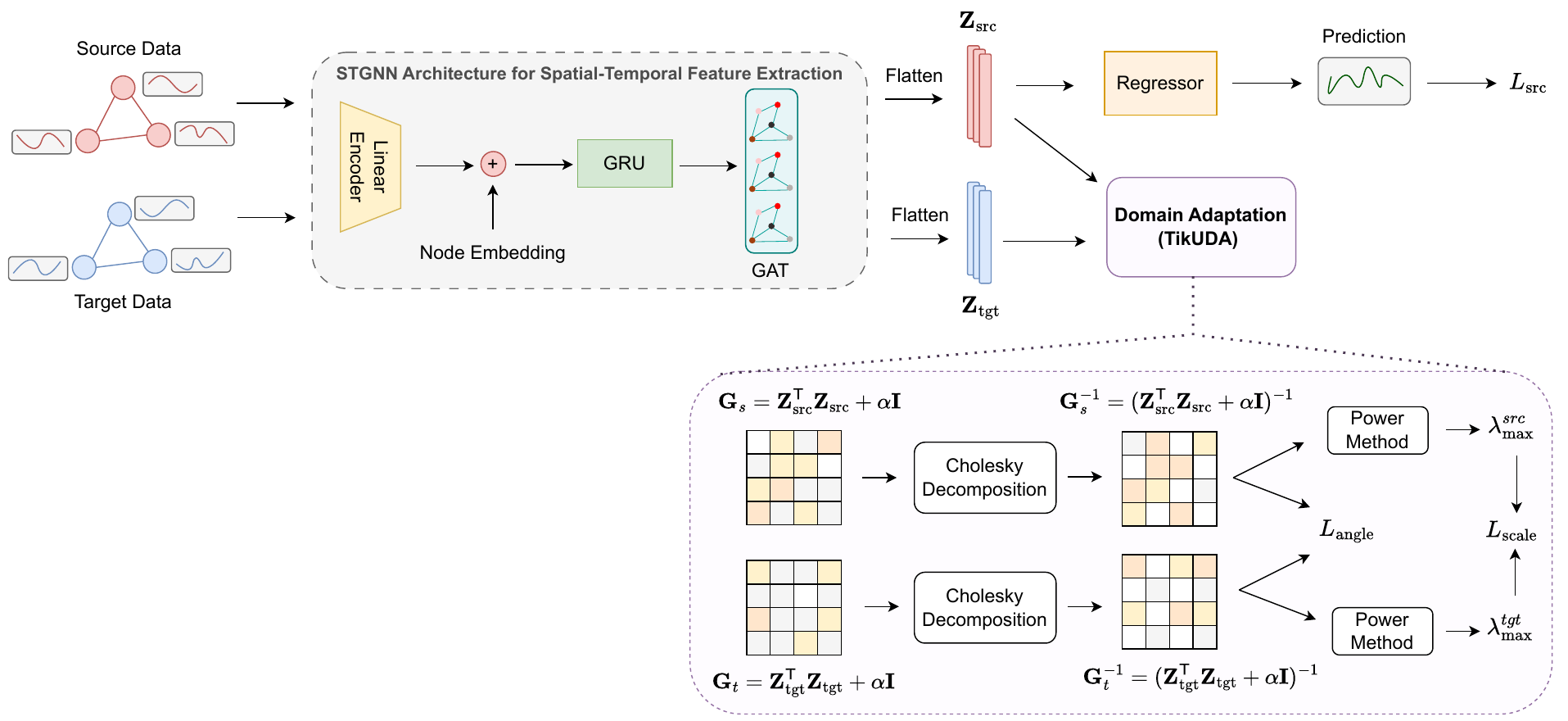}}
\caption{Overview of the proposed approach for domain-adaptive regression on spatial-temporal graph data. The model extracts temporal features using GRU and spatial dependencies via GAT, followed by a regression head for target prediction. The TikUDA module enables unsupervised domain adaptation by aligning the source and target distributions.}
\label{framework}
\end{figure*}

\noindent\textbf{Temporal Module of Feature Extractor :}
Temporal dependencies, also known as temporal patterns, can be effectively represented in both the time domain and the frequency domain \cite{jin2023survey}. To address these dependencies, we incorporate temporal modules before spatial modules, facilitating the modeling of complex spatial-temporal patterns. Our approach adopts recurrence-based methodologies, specifically leveraging the Gated Recurrent Unit (GRU), to capture and interpret the underlying temporal patterns inherent in the time series data.

\noindent\textbf{Spatial Module of Feature Extractor :}
We employ a series of message-passing operators to capture the spatial relationships among sensor measurements, iteratively updating representations at each layer \cite{hamilton2020graph}. The process of these iterative updates can be described as follows:
\begin{equation}
\mathbf{m}_{\mathcal{N}}^{(k)}(u) = \mathsf{AGG} \left\{ \left(\mathsf{MSG}^{k} (\mathbf{h}_{u}^{(k)}, \mathbf{h}_{v}^{(k)}), \, \forall v \in \mathcal{N}(u) \right) \right\},
\end{equation}
\begin{equation}
\mathbf{h}_{u}^{(k+1)} = \mathsf{UPDATE}^{(k)} \left( \mathbf{h}_{u}^{(k)}, \, \mathbf{m}_{\mathcal{N}}^{(k)}(u) \right),
\end{equation}
where $\mathbf{h}_{u}^{(k)}$ denotes the embedding of node $u$ at the $k$-th layer. The functions responsible for updating ($\mathsf{UPDATE}$) and messaging ($\mathsf{MSG}$) can be realized through any differentiable function, such as a multilayer perceptron (MLP). The aggregation operation ($\mathsf{AGG}$) represents a permutation-invariant aggregation operator. Here, $\mathcal{N}(u)$ denotes the set of neighboring nodes of node $u$. To improve the aggregation layer in GNNs, one can assign attention weights to each neighbor, regulating their influence during the aggregation process. In Graph Attention Networks (GAT) \cite{veličković2018graph}, attention weights are used to compute a weighted sum of neighboring node embeddings as follows:
\begin{equation}
\mathbf{m}_{\mathcal{N}(u)} = \sum_{v \in \mathcal{N}(u)} \alpha_{u,v} \mathbf{h}_v
\end{equation}
where $\alpha_{u,v}$ represents the attention weight assigned to neighbor $v \in \mathcal{N}(u)$ when aggregating information at node $u$. It can be defined as follows:
\begin{equation}
\alpha_{u,v} = \frac{\exp \left( \mathsf{LeakyReLU} \left( \mathbf{a}^\mathsf{T} \cdot [\mathbf{W}\mathbf{h}_u || \mathbf{W}\mathbf{h}_v] \right) \right)}{\sum_{v' \in \mathcal{N}(u)} \exp \left( \mathsf{LeakyReLU} \left( \mathbf{a}^\mathsf{T} \cdot [\mathbf{W}\mathbf{h}_u || \mathbf{W}\mathbf{h}_{v'}] \right) \right)}
\end{equation}
where $\mathbf{a}$ represents a trainable attention vector, $\mathbf{W}$ is a trainable matrix, and $||$ denotes the concatenation operation. For brevity, the $k$-th GAT layer can be represented compactly as:
\begin{equation}
\mathbf{H}^{(k)} = \mathsf{GAT}^{(k)}(\mathbf{H}^{(k-1)}, \mathbf{A}),
\end{equation}
where $\mathbf{A}$ is the adjacency matrix that represents the connections between nodes in the graph.

In terms of complexity, the GAT component scales as $\mathcal{O}(|\mathcal{V}| \cdot d^2 + |\mathcal{E}| \cdot d)$ per attention head, where $|\mathcal{V}|$ is the number of nodes, $|\mathcal{E}|$ is the number of edges, and $d$ is the feature dimension. This accounts for both the linear transformations of node features ($\mathcal{O}(|\mathcal{V}| \cdot d^2)$) and the attention coefficient computations over edges ($\mathcal{O}(|\mathcal{E}| \cdot d)$). For temporal modeling, GRU maintains a complexity of $\mathcal{O}(T \cdot N \cdot d^2)$, where $T$ is the sequence length, $N$ is the number of nodes, and $d$ is the hidden size, due to matrix operations across all nodes and time steps.

\subsection{Overall Framework and Summary}
\label{sec:overall}
The feature extraction architecture implemented in this study captures both temporal and spatial dependencies within the data, following a time-then-space approach as illustrated in Figure \ref{framework}. Initially, the input data undergoes an encoding process that projects the input features into a hidden space using a linear encoder. In IoT air quality sensor networks, sensors can display diverse characteristics that are interconnected in complex ways. For instance, in sensor networks including temperature, humidity, and low-cost gas sensors, temperature sensors may exhibit correlated (either similar or opposite) patterns of behavior, as might the humidity sensors. To accurately represent the different conditions affecting each sensor's behavior, we assign an embedding vector to each sensor to capture its unique characteristics. These embeddings are initially set to random values and then trained alongside the model, allowing them to adapt based on the data and interdependencies within the sensor network. 

Next, the data undergoes a temporal processing step using GRU to identify and extract temporal patterns. This is followed by a spatial processing step employing a GAT. If a graph structure is available (e.g., EEG signal), the GAT operates on it directly; otherwise (e.g., air quality), we construct a fully connected graph to capture spatial dependencies. The extracted features are subsequently inputted into a regressor to estimate the target values.

In UDA procedure, TikUDA is designed to align the inverse of the Gram matrix of features, which is perturbed by a scaled identity matrix referred to as the Tikhonov matrix. This perturbation ensures that the Gram matrix becomes full rank and invertible. A key aspect of the method is that the Tikhonov matrix is both symmetric and positive definite, which allows for the use of Cholesky decomposition—a more computationally efficient approach to matrix inversion compared to SVD, particularly for large matrices. Cholesky decomposition, specifically suited for symmetric and positive definite matrices, enhances numerical stability while maintaining the structural properties of the Tikhonov matrix. Additionally, TikUDA introduces novel similarity metrics (Haversine similarity) as regularization terms to improve subspaces' alignment between the source and target feature domains. Moreover, TikUDA focuses on scale alignment between source and target domains by aligning the largest principal eigenvalues of the inverse Tikhonov matrices, as the largest eigenvalue demonstrates the greatest influence on feature scaling. Instead of computing all eigenvalues through full SVD and then selecting the largest, TikUDA uses the power method to directly approximate the largest eigenvalue. This reduces the computational cost compared to full SVD, making the approach more efficient while ensuring effective alignment between source and target domains.

Finally, the training objective is to minimize both the prediction error on the source domain and the alignment of features between the source and target domains. The total loss function can be expressed as:

\begin{equation}
L_{\text{total}} = L_{\text{src}} + \gamma_{\text{angle}} L_{\text{angle}} + \gamma_{\text{scale}} L_{\text{scale}},
\label{finalloss}
\end{equation}

The loss term \( L_{\text{src}} \) represents the supervised loss (mean squared error) on the source domain, comparing the prediction with the ground truth. The terms \( L_{\text{angle}} \) and \( L_{\text{scale}} \) account for the angle and scale alignment with respect to the inverse Tikhonov matrix. The hyperparameters \( \gamma_{\text{angle}} \) and \( \gamma_{\text{scale}} \) control the influence of angle and scale alignment on the total loss.

\section{Experimental Setup} 
\label{sec:setup}

\subsection{Case Studies}
\begin{figure}
\centerline{\includegraphics[width=\linewidth]{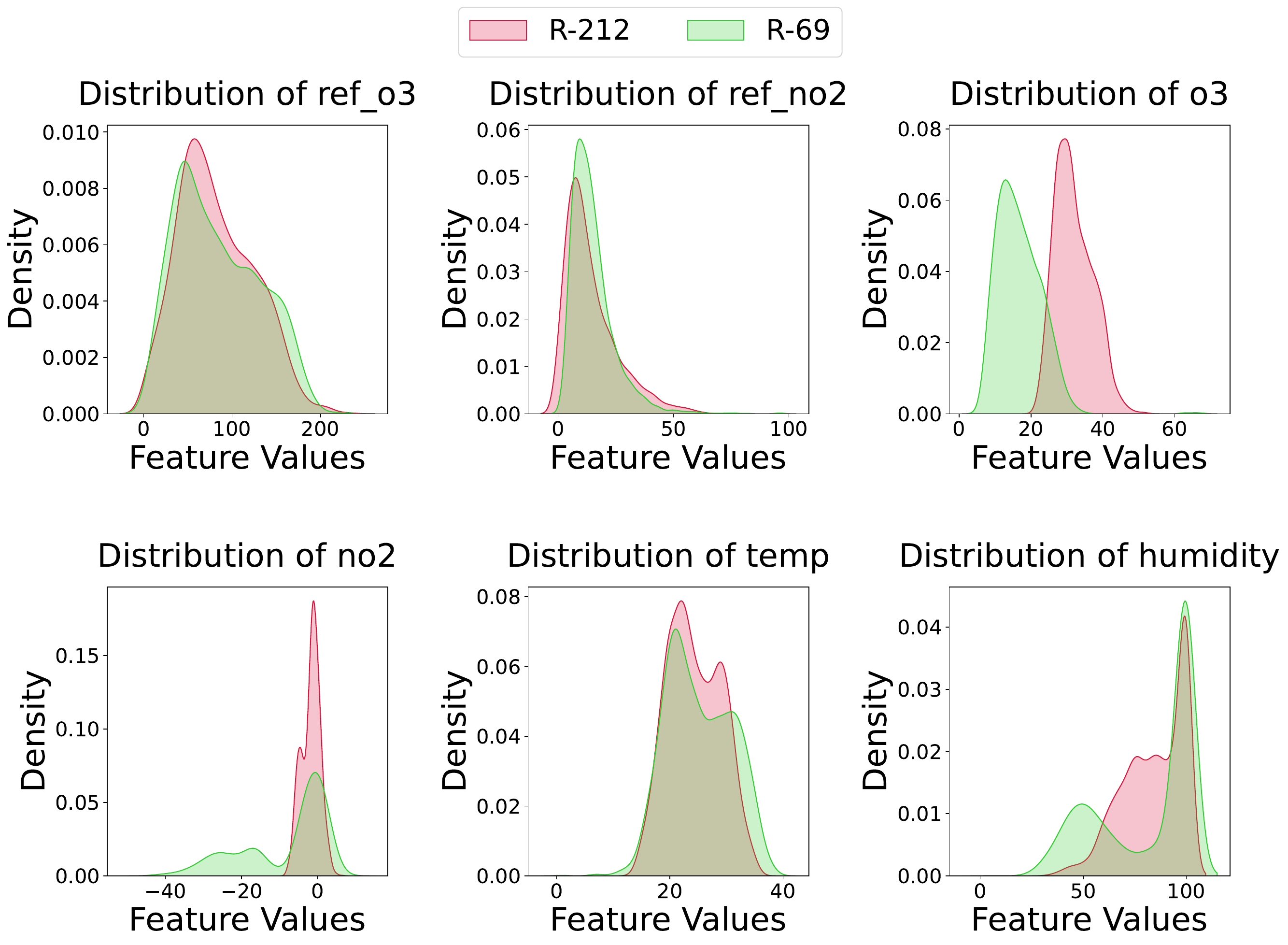}}
\caption{Distribution of features and labels between two datasets using Kernel Density Estimation (KDE) plots.}
\label{disribution}
\end{figure}

\begin{figure}
  \centering
  \begin{subfigure}[b]{\linewidth}
    \includegraphics[width=\linewidth]{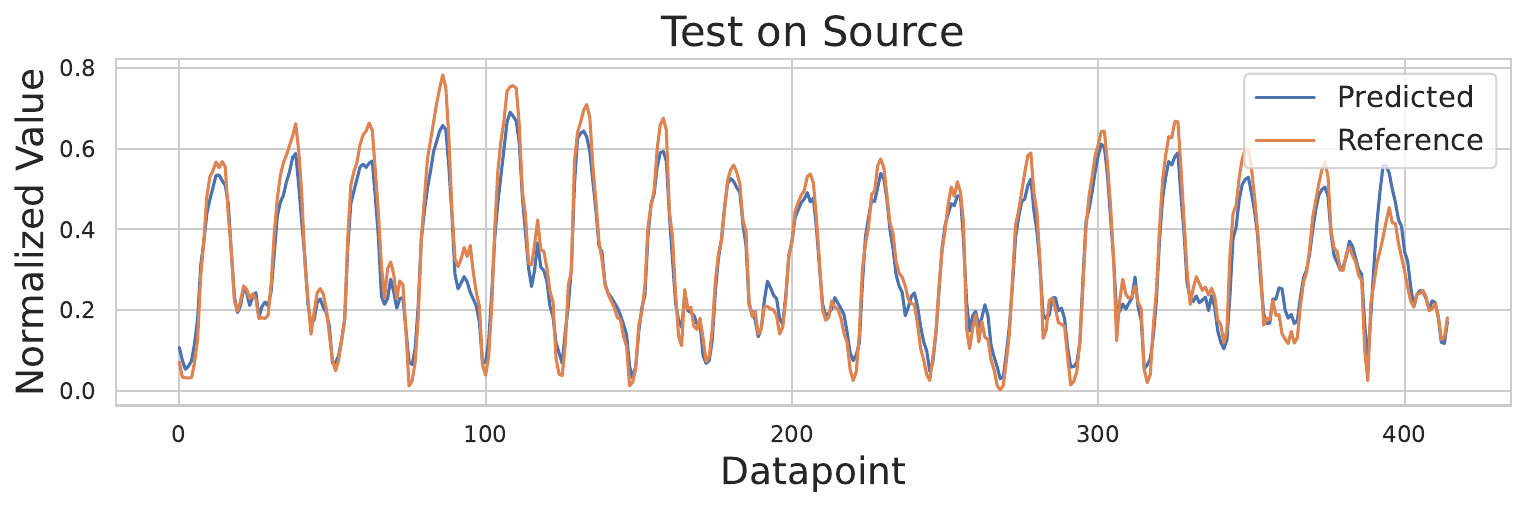}
  \end{subfigure}
  \hfill
  \begin{subfigure}[b]{\linewidth}
    \includegraphics[width=\linewidth]{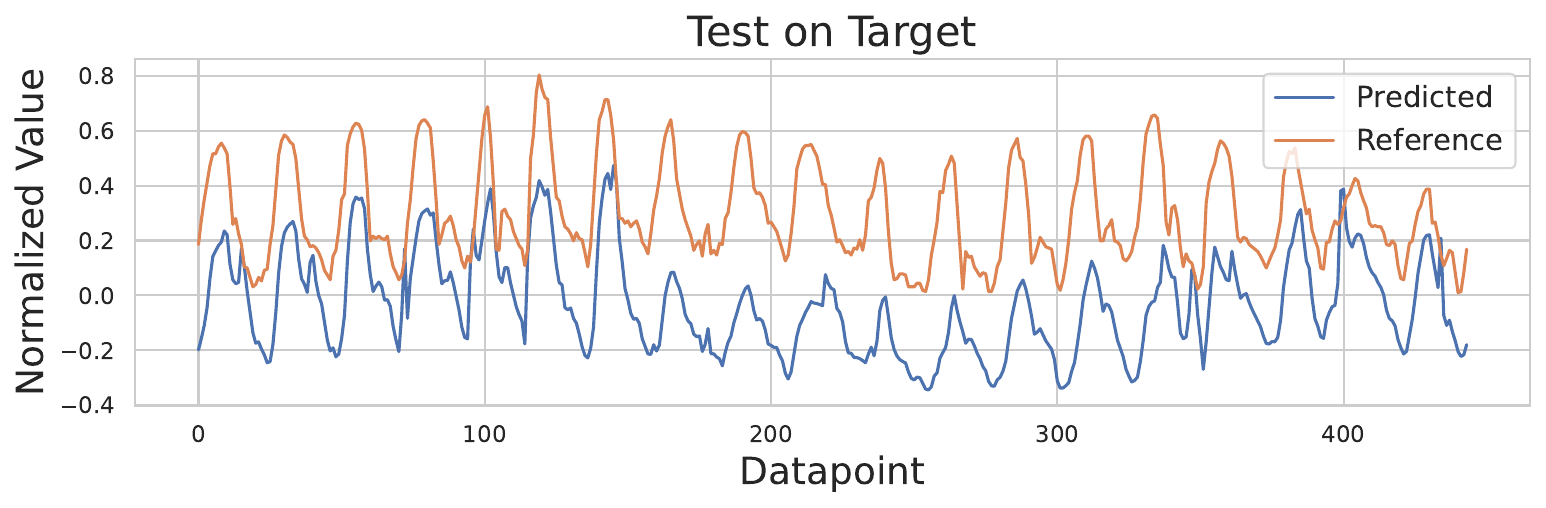}
    \label{fig:figure2}
  \end{subfigure}
  \caption{Impact of distribution shift between source and target data in different measurement locations on the model performance.}
  \label{fig:domainshift}
\end{figure}

\textbf{Air Quality Sensor Networks:} We assess our proposed method using two publicly available real-world datasets from IoT air quality sensor networks \cite{ripoll2019testing, barcelo2021h2020}. The networks are located in Monte Cucco, an urban area in the province of Piacenza (Italy), designated as R-N69, and in Osio Sotto in Bergamo (Italy), designated as R-N212. Both networks were installed near reference stations operated by governmental agencies using high-precision instrumentation, providing a reliable ground truth for evaluation. Each network is equipped with a low-cost O\textsubscript 3 sensor, NO\textsubscript 2 sensor, temperature sensor, and relative humidity sensor. A detailed summary of the datasets is provided in Table \ref{tab:DatasetConfig} \cite{allka}. These two networks are separated by a considerable distance, leading to a distribution shift in sensor measurements, as illustrated in Figure \ref{disribution}. We evaluate the two datasets in both configurations: first using one dataset as the source and the other as the target, then reversing their roles. In Figure \ref{fig:domainshift}, we illustrate the effect of distribution shift on model performance in our experimental setup. As depicted, when the model is trained and deployed at the same measurement location, the performance remains competitive due to the fact that the distribution of the features remains constant. However, when the model is deployed at a different measurement location, a significant drop in performance is observed. This drop is attributed to the distribution shift between the source and target locations. We imitate a practical scenario where reliable ground truth measurements are not available at the target location. The task involves training a model on one labeled dataset (source) to predict reliable O\textsubscript3 and NO\textsubscript2 values at the target location. 

\textbf{Electroencephalography (EEG) Case Study}: EEG signals are widely used in neuroscience and biomedical research for their ability to  non-invasively capture the brain's electrical activity. We selected an EEG dataset for this study due to its inherent complexity and variability, both within and across individuals. This  natural inter-subject variability presents  a challenging yet realistic testbed  for evaluating domain adaptation methods. Effective cross-subject knowledge transfer is essential in applications such as brain-computer interfaces and neurological disorder diagnostics, where models must generalize across individuals with distinct  brainwave patterns -- making EEG an ideal benchmark for our proposed TikUDA method.

\begin{figure}
\centering
\includegraphics[width=0.65\linewidth]{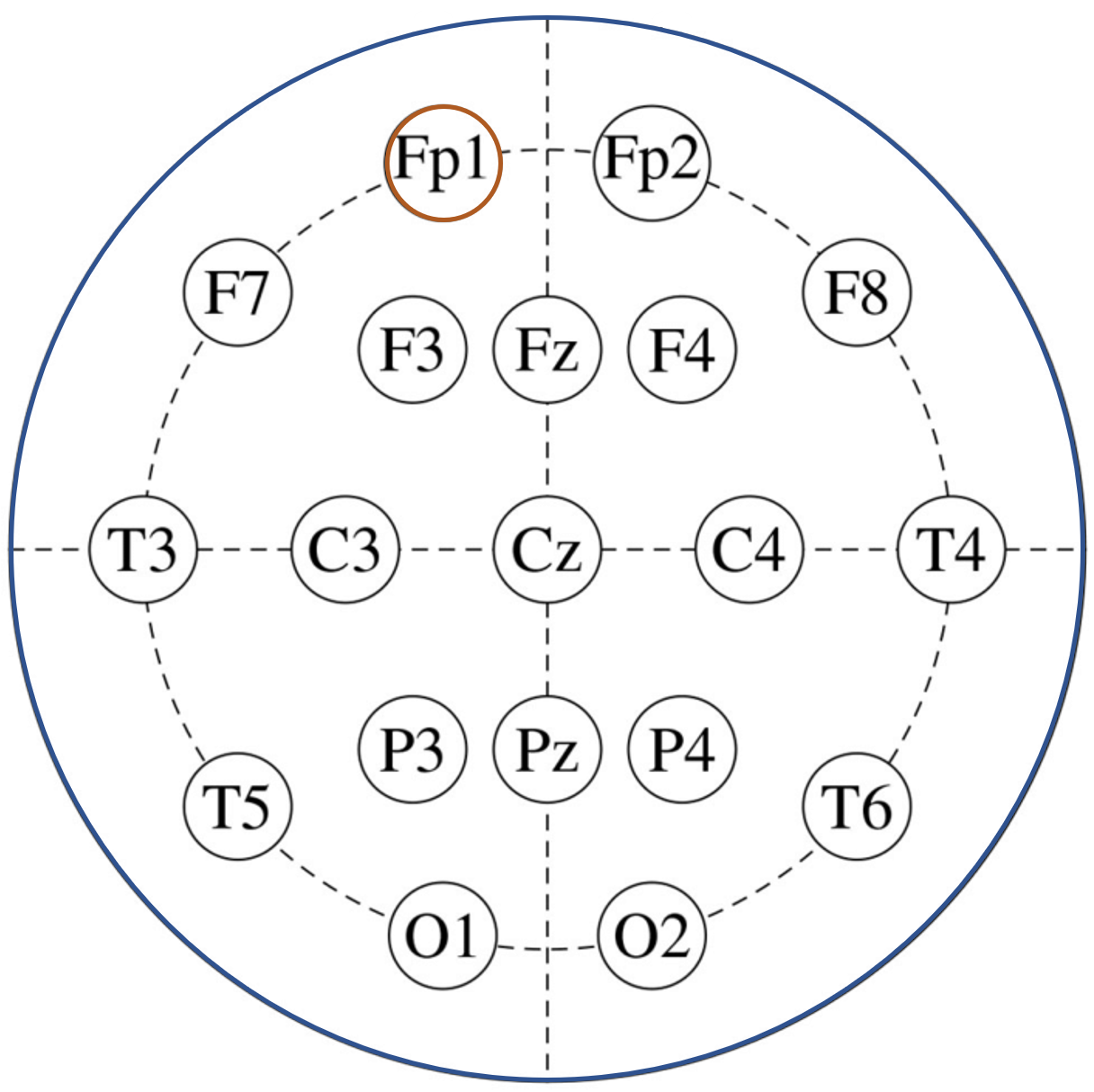}
\caption{EEG setup with 19 channels.}
\label{eeg_rev}
\end{figure}

The EEG data, obtained from \cite{zyma2019electroencephalograms}, were recorded using a Neurocom 23-channel EEG system ( XAI-MEDICA, Ukraine)  in a monopolar configuration. Electrodes were positioned  according to the international 10/20 system and referenced to interconnected ear electrodes. Data preprocessing included a 30 Hz high-pass filter and a 50 Hz notch filter to remove power-line noise, as defined in previous studies \cite{zyma2019electroencephalograms}. Independent Component Analysis (ICA) was applied  to remove artifacts caused by eye movements, muscle activity, and cardiac pulsations.  Only artifact-free segments of 60 seconds were used for analysis. During each trial, participants engaged in a cognitive task involving the serial subtraction of a two-digit number from  a four-digit number, presented  orally at the start. For the regression task, we focused on reconstructing the FP1 channel using the remaining active channels: Fp1, Fp2, F3, F4, F7, F8, T3, T4, C3, C4, T5, T6, P3, P4, O1, O2, Fz, Cz, and Pz, as illustrated in Fig. \ref{eeg_rev}  In the domain adaptation setup, we selected two subjects:  Subject 33 (a 17-year-old male) as the source and Subject 00 (a 21-year-old female) as the target. Due to a malfunctioning EEG cap, the FP1 channel was not properly recorded for for Subject 00, simulating a real-world scenario of sensor failure  or noise. The task, therefore, involves leveraging  data from other channels and the source subject to  reconstruct  missing measurements for the target subject -- testing  the model’s  robustness to both missing data and inter-subject domain shifts.

\begin{table*}[ht!]
  \centering
  \caption{Summary of the Air Quality Datasets Used in the Study,}
  \resizebox{\linewidth}{!}{
  \begin{tabular}{cccccc}
    \toprule
    Dataset Name & Sensor Arrays (outputs are \textbf{bolded}) & Sensor Type & Location & Samples & Timeline \\
    \toprule
    R-69 & \makecell{\textbf{O\textsubscript{3}, NO\textsubscript{2}}, O\textsubscript{3} (low-cost), NO\textsubscript{2} (low-cost), \\ Temperature, Relative Humidity} & \makecell{OX-B431/NO2-B43F \\ DHT1-Grove} & Monte Cucco & \makecell{1 per hour} & \makecell{21/06/2018 - 25/09/2018} \\
    \midrule
    R-212 & \makecell{\textbf{O\textsubscript{3}, NO\textsubscript{2}}, O\textsubscript{3} (low-cost), NO\textsubscript{2} (low-cost), \\ Temperature, Relative Humidity}& \makecell{OX-B431/NO2-B43F \\ DHT1-Grove} & Osio Sotto & \makecell{1 per hour} & \makecell{27/06/2018 - 24/09/2018} \\
    \bottomrule
  \end{tabular}
  }
  \label{tab:DatasetConfig}
\end{table*}

\subsection{Data Preprocessing}
To prepare the data, each sensor measurement was scaled to a range of $[0, 1]$ using a min-max normalization approach. The scaling parameters were fitted on the source domain data for each adaptation task and then applied to the target domain data. All data from both domains were used in training for each adaptation task, and no train-test splitting was used. Sensor measurements were segmented into samples using a sliding window approach with a window size of 16 and a step size of 1.

\subsection{Evaluation Metrics}
We evaluate and compare the performance of the proposed framework using two commonly used regression evaluation metrics: root-mean-square error (RMSE) and mean absolute error (MAE).

To provide a more comprehensive analysis, we evaluate the methods using both normalized and unnormalized labels. 
Normalized labels allow for results to be compared on a common scale, which is particularly useful when working with diverse datasets. Conversely, unnormalized labels retain the original scale of the data, providing insights into real-world applications.

\subsection{Comparison Methods}

In this subsection, we present a comparative analysis of our proposed TikUDA algorithm against various UDA methods that address the challenges posed by distribution shifts between source and target domains. Each method employs a unique strategy to mitigate these distribution shifts, providing a thorough and comprehensive comparison.

The following methods are considered for the comparison:
\begin{itemize}
  \item \textbf{Source-Only:} A baseline method where a model is trained exclusively on the source domain data and then directly applied to the target domain data. This approach does not involve any domain adaptation techniques, potentially leading to degraded performance when there is a distribution shift between the source and target domains. This approach usually serves as the lower bound of the performance.

  \item \textbf{Maximum Mean Discrepancy (MMD):} A distance-based domain adaptation method that assesses the difference between two empirical probability distributions \cite{long2015learning}. It embeds the distributions in a reproducing kernel Hilbert space to measure the distance between them.

  \item \textbf{Adaptive Batch Normalization (AdaBN):} AdaBN \cite{li2018adaptive} is a domain adaptation method that adjusts the batch normalization parameters of a trained model based on the target domain data. Unlike other methods that focus on learning domain-invariant features, AdaBN modifies the batch normalization statistics to fit the target domain. The model parameters are trained solely on the source domain data, with target domain data used only to update the normalization statistics. This requires the feature extractor to include batch normalization layers to implement AdaBN.

  \item \textbf{Correlation Alignment (CORAL):} CORAL \cite{sun2016deep} aligns the second-order statistics (mean and covariance) of the source and target domain features. This method distinguishes itself from subspace manifold approaches by aligning the original feature distributions of the source and target domains, rather than focusing on the bases of lower-dimensional subspaces.

  \item \textbf{Empirical Risk Minimization with Nuclear Norm Regularization (ERM-NU):} ERM-NU \cite{shi2024domain} combines empirical risk minimization with nuclear norm regularization. This method addresses low-rank matrix completion and recovery problems using nuclear norm minimization. ERM-NU aims to obtain domain-invariant features. By minimizing the nuclear norm of the features, ERM-NU seeks to reduce the rank and thus enhance the alignment between domains.

  \item \textbf{Adaptive Graph Convolutional Network (AdaGCN):} AdaGCN \cite{dai2022graph} leverages adversarial domain adaptation and graph convolution to transfer label information from labeled source data to unlabeled target data. It uses an adversarial domain adaptation component that mitigates the distribution shift between the source and target domains.

  \item \textbf{Aligning the Inverse GRAM Matrices (DARE-GRAM):} DARE-GRAM aligns the inverse GRAM matrices of the source and target domains to mitigate distribution shifts. This alignment strategy aims to improve model performance on the target domain by reducing discrepancies between the domain distributions.
\end{itemize}

\subsection{Implementation Details}
In this study, all compared methods share the same base architecture, except for AdaBN, which includes an additional batch normalization layer after the GAT layer to ensure a fair comparison. To determine the number of layers and the hidden dimension of the feature extractor, we split the source domain dataset into training and testing sets to perform model selection based on the labeled source data. This approach ensures that the feature extractor can effectively extract the necessary informative features for multisensor fusion and calibration tasks. The feature extractor consists of four GRU layers and one GAT layer, with all layers (linear encoder, GRU, and GAT) set to a hidden dimension of 16. The remaining architecture includes a regressor, comprising one fully connected layer to predict the final output value. The model is updated using the Adam optimizer \cite{kingma} with a learning rate of \(3 \times 10^{-4}\) and batch sizes of 64, over 150 epochs. For each training step, one batch of source and target domain data is processed.

In UDA, the absence of labeled data for the target domain makes it impossible to utilize a validation dataset for hyperparameter tuning. Consequently, an alternative approach must be adopted. We employed a heuristic method proposed in \cite{sun2019unsupervised}, which addresses this challenge. Our observations indicated that if the trade-off alignment regularization is set too high for all methods, it hinders the model's ability to learn effectively. Conversely, if it is too low, there is insufficient adaptation to the target domain. To tune the hyperparameters, we conducted a search over specific values and tracked the normalized summation of the source loss and the trade-off alignment loss. This process was applied to all methods to ensure a fair comparison. In our approach, the hyperparameters \(\gamma_{\text{angle}}\) and \(\gamma_{\text{scale}}\), as defined in equation (\ref{finalloss}), were set to \(\gamma_{\text{angle}} = 10^{-2} \lambda\) and \(\gamma_{\text{scale}} = 10^{-3} \lambda\), respectively for the first scenario (air quality case study). Here, \(\lambda = \frac{2}{1 + \exp(-10p)} - 1\) and \(p\) represents the progress of training iterations, allowing \(\lambda\) to range from 0 to 1. Consequently, \(\gamma_{\text{angle}}\) and \(\gamma_{\text{scale}}\) vary from 0 to \(10^{-2}\) and \(10^{-3}\), respectively. This follows a similar approach to \cite{ganin2015unsupervised, nejjar2024uncertainty}. For EEG data, hyperparameters were set to \(\gamma_{\text{angle}} = 5 \times 10^{-2} \lambda\) and \(\gamma_{\text{scale}} = 2.5 \times 10^{-3} \lambda\).

\section{Results and Discussion}
\label{sec:result}

\begin{table*}[ht!]
  \centering
  \caption{RMSE Results for Four Domain Adaptation Scenarios}
  \resizebox{\linewidth}{!}{
  \begin{tabular}{lcc|cc|cc|cc|cc}
    \toprule
    Scenario & \multicolumn{2}{c|}{R-212 $\rightarrow$ R-69 (O\textsubscript3)} & \multicolumn{2}{c|}{R-212 $\rightarrow$ R-69 (NO\textsubscript2)} & \multicolumn{2}{c|}{R-69 $\rightarrow$ R-212 (O\textsubscript3)} & \multicolumn{2}{c|}{R-69 $\rightarrow$ R-212 (NO\textsubscript2)} & \multicolumn{1}{c}{Avg.} \\
    \midrule
    \textbf{Method} & \textbf{Norm.} \(\downarrow\) & \textbf{Actual} \(\downarrow\) & \textbf{Norm.} \(\downarrow\) & \textbf{Actual} \(\downarrow\) & \textbf{Norm.} \(\downarrow\) & \textbf{Actual} \(\downarrow\) & \textbf{Norm.} \(\downarrow\) & \textbf{Actual} \(\downarrow\) & \textbf{Norm.} \(\downarrow\) \\
    \midrule
    Source-Only & 0.255 & 59.40 & 0.183 & 11.88 & 0.217 & 48.44 & 0.117 & 11.11 & 0.193 \\
    MMD \cite{long2015learning} & 0.159 & 37.13 & 0.139 & 9.03 & 0.118 & 26.25 & 0.105 & 9.99 & 0.130  \\
    AdaBN \cite{li2018adaptive} & 0.254 & 59.39 & 0.289 & 18.67 & 0.118 & 26.29 & 0.106 & 10.05 & 0.192 \\
    CORAL \cite{sun2016deep} & 0.205 & 47.71 & 0.158 & 10.27 & 0.175 & 39.03 & 0.104 & 9.89 & 0.161 \\
    ERM-NU \cite{shi2024domain} \ & 0.184 & 43.19 & 0.146 & 9.48 & 0.124 & 27.62 & 0.102 & 9.67 & 0.139  \\
    AdaGCN  \cite{dai2022graph} & 0.103 & 23.96 & 0.142 & 9.17 & 0.114 & 25.53 & 0.104 & 9.92 & 0.116  \\
    DARE-GRAM \cite{nejjar2023dare} & \underline{0.094} & \underline{22.01} & \textbf{0.133} & \textbf{8.60} & \underline{0.102} & \underline{22.37} & \textbf{0.098} & \textbf{9.35} & \underline{0.107}  \\
    TikUDA & \textbf{0.087} & \textbf{20.35} & \underline{0.137} & \underline{8.88} & \textbf{0.097} & \textbf{21.87} & \underline{0.099} & \underline{9.37} & \textbf{0.105} \\
    \bottomrule
  \end{tabular}
  }
  \label{tab:rmse_results}
\end{table*}

\begin{table*}[ht!]
  \centering
  \caption{MAE Results for Four Domain Adaptation Scenarios}
  
  \resizebox{\linewidth}{!}{
  \begin{tabular}{lcc|cc|cc|cc|cc}
    \toprule
    Scenario & \multicolumn{2}{c|}{R-212 $\rightarrow$ R-69 (O\textsubscript3)} & \multicolumn{2}{c|}{R-212 $\rightarrow$ R-69 (NO\textsubscript2)} & \multicolumn{2}{c|}{R-69 $\rightarrow$ R-212 (O\textsubscript3)} & \multicolumn{2}{c|}{R-69 $\rightarrow$ R-212 (NO\textsubscript2)} & \multicolumn{1}{c}{Avg.} \\
    
        \midrule
    \textbf{Method} & \textbf{Norm.} \(\downarrow\) & \textbf{Actual} \(\downarrow\) & \textbf{Norm.} \(\downarrow\) & \textbf{Actual} \(\downarrow\) & \textbf{Norm.} \(\downarrow\) & \textbf{Actual} \(\downarrow\) & \textbf{Norm.} \(\downarrow\) & \textbf{Actual} \(\downarrow\) & \textbf{Norm.} \(\downarrow\) \\
    \midrule
    Source-Only & 0.221 & 51.47 & 0.135 & 8.69 & 0.194 & 43.35 & 0.092 & 8.79 & 0.161 \\
    MMD \cite{long2015learning} & 0.129 & 30.21 & 0.097 & 6.30 & 0.090 & 20.16 & 0.080 & 7.62 & 0.099  \\
    AdaBN \cite{li2018adaptive} & 0.209 & 48.95 & 0.241 & 15.60 & 0.094 & 20.93 & 0.082 & 7.76 & 0.157  \\
    CORAL \cite{sun2016deep} & 0.168 & 39.28 & 0.115 & 7.41 & 0.148 & 32.95 & 0.079 & 7.51 & 0.128  \\
    ERM-NU \cite{shi2024domain} \ & 0.159 & 37.29 & 0.112 & 7.25 & 0.099 & 22.04 & \underline{0.073} & \underline{6.92} & 0.111  \\
    AdaGCN  \cite{dai2022graph} & 0.083 & 19.38 & 0.098 & 6.31 & 0.090 & 20.11 & 0.077 & 7.39 & 0.087  \\
    DARE-GRAM \cite{nejjar2023dare} & \underline{0.077} & \underline{17.83} & \underline{0.095} & \underline{6.11} & \underline{0.081} & \underline{17.66} & \underline{0.073} & 6.94 & \underline{0.082}  \\
    TikUDA & \textbf{0.070} & \textbf{16.38} & \textbf{0.093} & \textbf{6.02} & \textbf{0.076} & \textbf{17.17} & \textbf{0.072} & \textbf{6.79} & \textbf{0.078} \\
    \bottomrule
  \end{tabular}
  }
  \label{tab:mae_results}
\end{table*}

In this section, first, we assess the effectiveness of all compared methods based on their performance using RMSE and MAE metrics for air quality dataset, as shown in Tables \ref{tab:rmse_results} and \ref{tab:mae_results}, respectively. The source-only model reveals that the prediction error for the O\textsubscript{3} is more than four times larger than for the NO\textsubscript{2} task in both transfer scenarios, indicating a significant domain shift in O\textsubscript{3} estimation. Specifically, for the O\textsubscript{3} task, our proposed method significantly improves RMSE from 59.40 (source-only model) to 20.35 (TikUDA) for the transfer task \textit{`R-212 to R-69'}, and from 48.44 (source-only model) to 21.87 (TikUDA) for the transfer task \textit{`R-69 to R-212'.}
Overall, TikUDA demonstrates the best average performance compared to other methods across all adaptation and transfer scenarios, achieving an average RMSE of 0.105 and an average MAE of 0.078, outperforming all other methods. DARE-GRAM shows the second-best performance with an average RMSE of 0.107 and an average MAE of 0.082, indicating that the proposed haversine distance function effectively helps align the source and target domains. Other methods such as MMD, CORAL, ERM-NU, and AdaGCN also showed improvements in RMSE and MAE compared to the baseline, as indicated in Tables \ref{tab:rmse_results} and \ref{tab:mae_results}. However, they did not achieve the same level of performance, as they are not primarily designed for regression tasks like DARE-GRAM and TikUDA.

For the EEG dataset, we employed  the same set of domain adaptation  methods as in the air quality task, using an identical STGNN architecture. This consistency ensures a fair and direct comparison across domains. The results in Table \ref{tab:rmse_results_EEG} show  that while several methods perform well in the domain adaptation scenario (S33 $\rightarrow$ S00), TikUDA  consistently achieves the best performance. Specifically, although  AdaGCN and ERM-NU achieve competitive results, TikUDA surpasses both, achieving the lowest RMSE of 0.178 and the lowest MAE of 0.130. These results  demonstrate TikUDA’s effectiveness and robustness  in EEG-based domain adaptation, particularly in handling inter-subject variability and missing data. Furthermore, they underscore  TiKUDA's broad applicability to  UDA regression tasks across diverse  domains.

\begin{table}[ht!]
  \centering
    \caption{Results for Domain Adaptation Scenario (S33 $\rightarrow$ S00) on EEG Signals}
  \resizebox{\linewidth}{!}{
  \begin{tabular}{lcc|cccc}
    \toprule
    
    \textbf{Method} & \textbf{RMSE (Norm.) \(\downarrow\)} & \textbf{RMSE (Actual) \(\downarrow\)} & \textbf{MAE (Norm.) \(\downarrow\)} & \textbf{MAE (Actual) \(\downarrow\)} & \\
    \midrule 
    Source-Only & 0.327 & 21.96 & 0.253 & 17.01 \\
    MMD & 0.182 & 12.24 & 0.134 & 8.99\\
    AdaBN & 0.294 & 19.78 & 0.225 & 15.18 \\
    CORAL & 0.318 & 21.44 & 0.246 & 16.59 \\
    ERM-NU & \underline{0.180} & 12.12 & \underline{0.132} & \underline{8.86}  \\
    AdaGCN & \underline{0.180} & \underline{12.11} & 0.137 & 9.22 \\
    DARE-GRAM & 0.205 & 13.82 & 0.154 & 10.40 \\
    TikUDA & \textbf{0.178} & \textbf{12.03} & \textbf{0.130} & \textbf{8.76} \\
    \bottomrule
  \end{tabular}
  }

  \label{tab:rmse_results_EEG}
\end{table}

\begin{figure*}[ht]
  \centering

  \begin{subfigure}[b]{0.24\textwidth}
    \includegraphics[width=\linewidth]{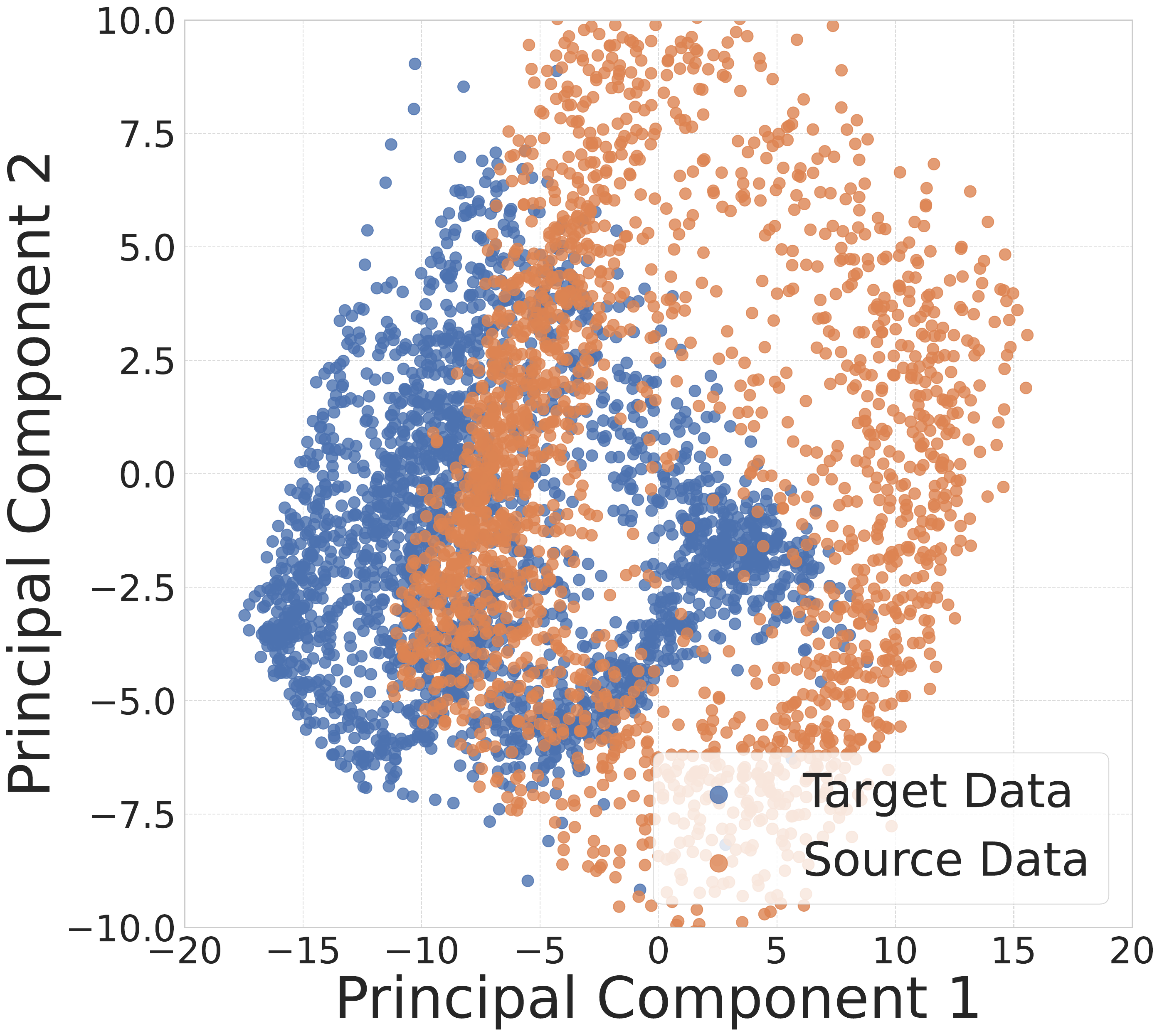}
    \caption{\shortstack{Baseline\\Energy Distance = 0.3877}}
  \end{subfigure}
  \hfill
  \begin{subfigure}[b]{0.24\textwidth}
    \includegraphics[width=\linewidth]{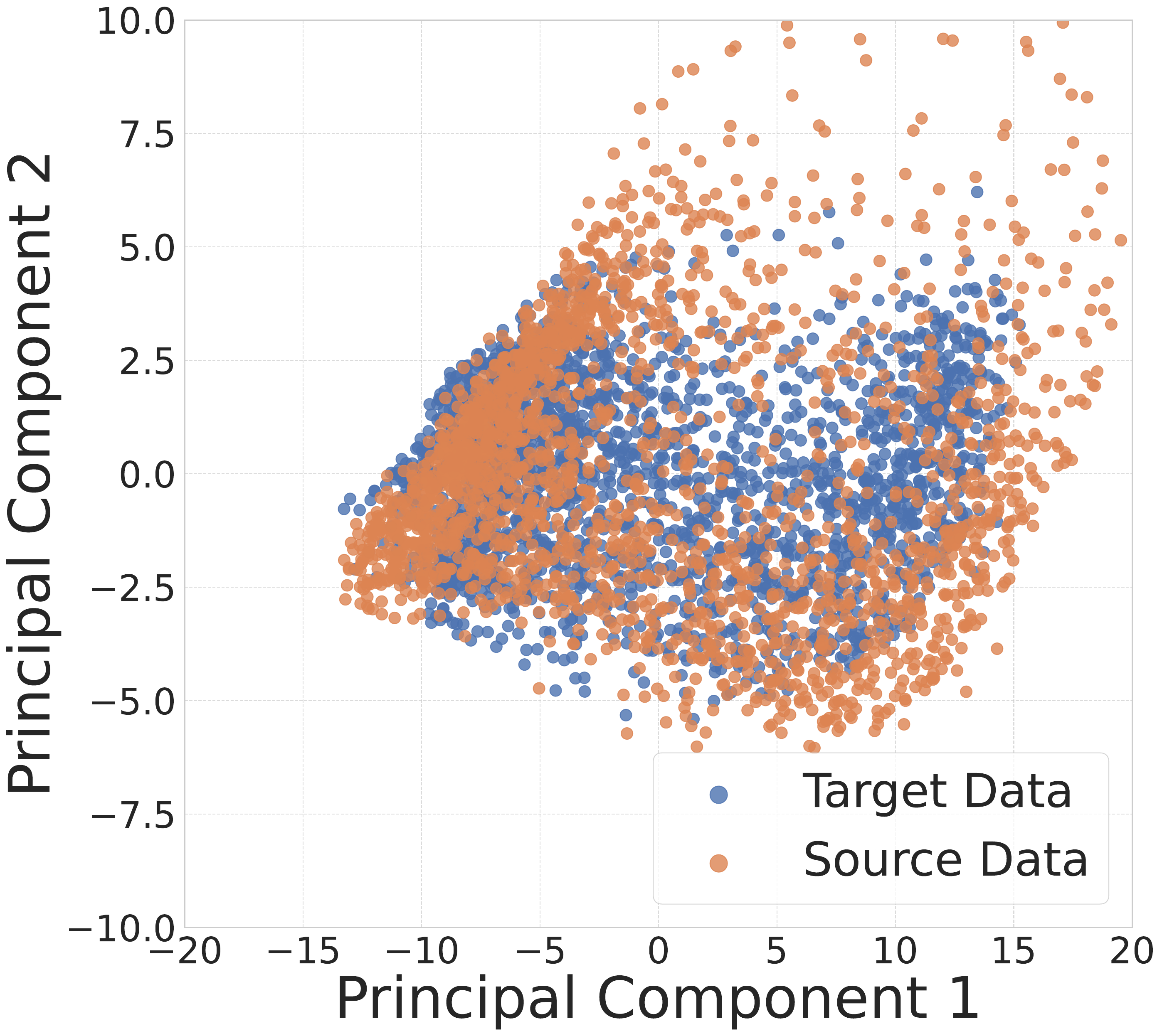}
    \caption{\shortstack{AdaGCN\\Energy Distance = 0.0790}}
  \end{subfigure}
  \hfill
  \begin{subfigure}[b]{0.24\textwidth}
    \includegraphics[width=\linewidth]{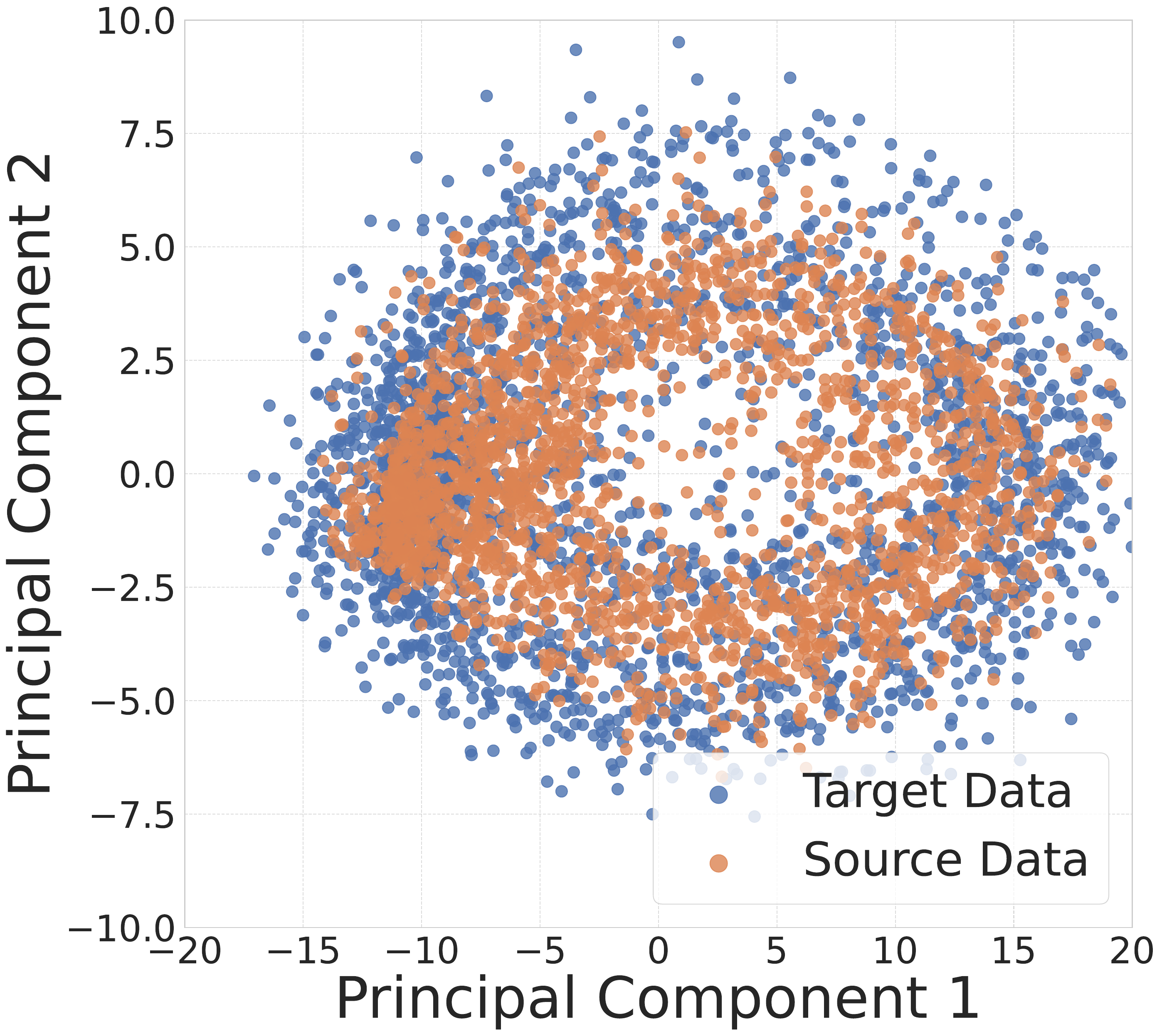}
    \caption{\shortstack{DARE-GRAM\\Energy Distance = 0.0128}}
  \end{subfigure}
  \hfill
  \begin{subfigure}[b]{0.24\textwidth}
    \includegraphics[width=\linewidth]{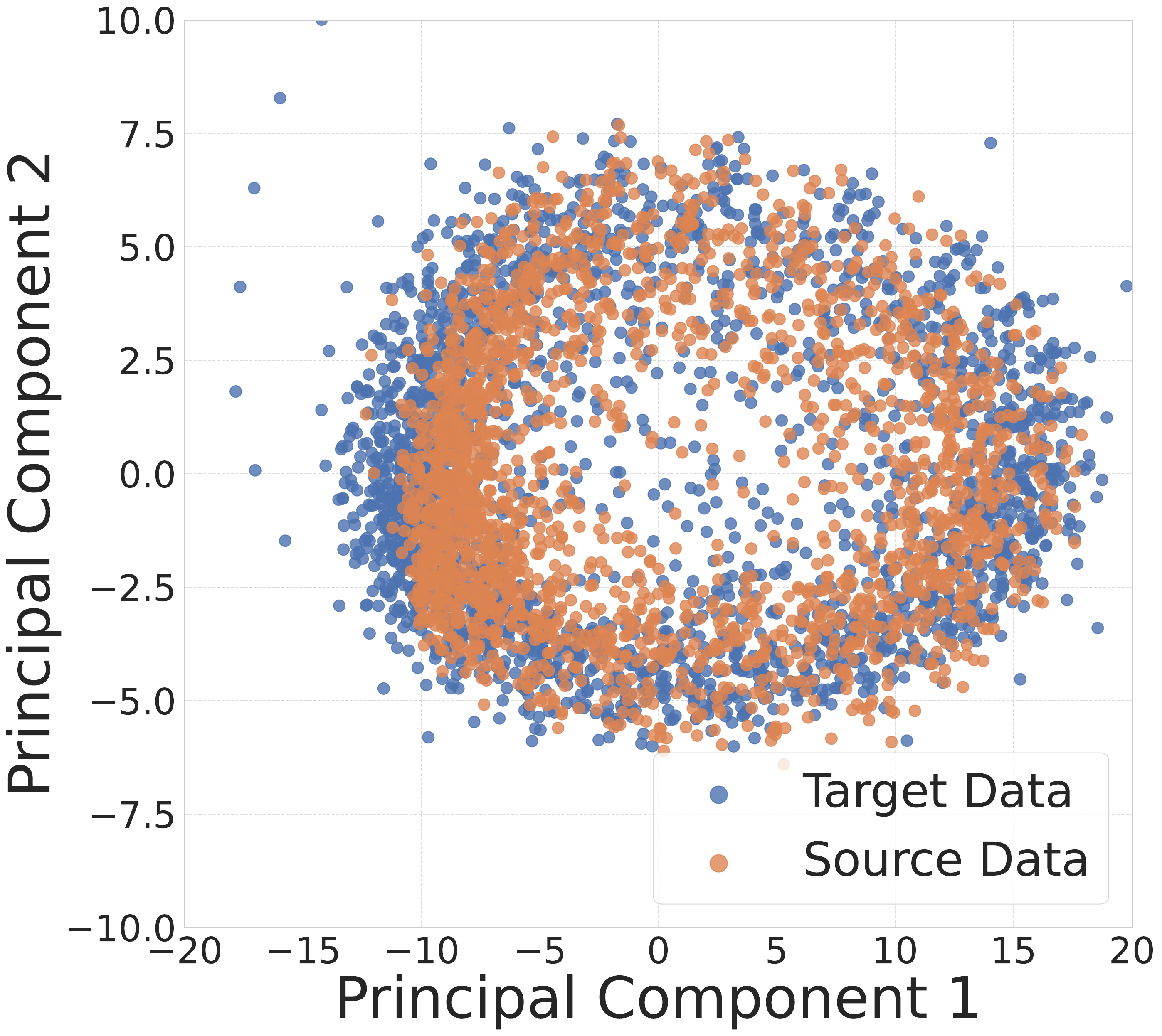}
    \caption{\shortstack{TikUDA\\Energy Distance = 0.0031}}
  \end{subfigure}

  \caption{PCA visualization of feature alignment for the R-212  $\rightarrow$ R-69 (O\textsubscript{3}) adaptation task. The baseline shows poor alignment between domains, while the proposed method achieves close overlap. Energy distance confirms improved alignment, with TikUDA yielding the lowest distance.}
  \label{visualization}
\end{figure*}

Additionally, we evaluate the impact of our approach on the embeddings generated during the alignment process for the adaptation scenario of R-212 $\rightarrow$ R-69 (O\textsubscript{3}). We applied principal component analysis (PCA) to the embedded features and plotted the first two principal components to create two-dimensional visualizations, as depicted in Figure \ref{visualization}. 
The PCA plot highlights a significant discrepancy between the source and target data distributions when using the source-only model. In contrast, our proposed method achieves a better alignment of data points from both domains, demonstrating its effectiveness in mitigating distribution shifts in the feature space. To quantitatively support this visual evidence, we compute the energy distance \cite{rizzo2016energy} between the source and target feature distributions for three competitive methods. The baseline model exhibits a high energy distance of 0.3877, indicating substantial domain misalignment. In comparison,  domain adaptation methods such as AdaGCN and DARE-GRAM reduce this distance to 0.0790 and 0.0128, respectively. Our proposed method achieves the lowest energy distance of 0.0031, indicating the most effective alignment of feature distributions and  the strongest reduction in domain shift among all evaluated approaches.

Furthermore, we compare the time complexity of our proposed algorithm, TikUDA, with DARE-GRAM, as both not only achieved the best performance in our experiments but also share similar theoretical foundations based on subspace methods for regression tasks. This makes a direct comparison of their running times both relevant and meaningful. Both DARE-GRAM and TikUDA exhibit similar asymptotic computational complexities, approximately \(\mathcal{O}(p^3)\), but differ significantly in their practical execution. DARE-GRAM relies on full SVD and the computation of the Moore-Penrose pseudo-inverse, which leads to higher computational overhead. Specifically, calculating the Gram matrix for both methods requires \(\mathcal{O}(bp^2)\), where \(b\) is the batch size and \(p\) is the feature dimension. The SVD step further introduces an \(\mathcal{O}(p^3)\) complexity, and computing the pseudo-inverse also contributes an additional \(\mathcal{O}(p^3)\), making DARE-GRAM computationally expensive. In contrast, TikUDA utilizes power iteration to approximate the largest singular value, avoiding the need for full SVD. Each iteration of the power method takes \(\mathcal{O}(p^2)\). Moreover, TikUDA leverages Cholesky decomposition, which has a complexity of \(\mathcal{O}(p^3/3)\), for matrix inversion, offering a more efficient alternative to SVD-based inversion. While both methods have similar theoretical upper bounds, the use of power iteration and Cholesky decomposition in TikUDA significantly reduce the constant factor in the complexity, leading to faster performance in practice compared to DARE-GRAM. Consequently, TikUDA is expected to perform more efficiently, especially for large-scale data, where full SVD can be prohibitively slow.

\begin{figure}
\centering
\includegraphics[width=\linewidth]{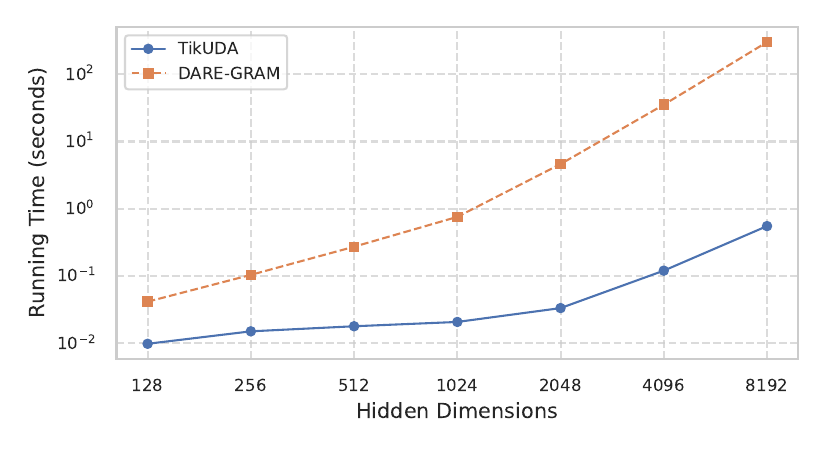}
\caption{Comparison of running times on training iteration for TikUDA and DARE-GRAM across varying hidden dimensions ($128$ to $8192$).}
\label{time}
\end{figure}

For comparing practical time complexity, Figure \ref{time} illustrates the comparison of running times between DARE-GRAM and TikUDA across different sizes of hidden dimensions during each training iteration conducted on NVIDIA GeForce RTX 2080 Ti GPU. As the hidden dimensions increase, the computation time for DARE-GRAM escalates significantly, whereas TikUDA maintains a much faster performance in high-dimensional settings. This efficiency is particularly beneficial for graph-structured data, where the number of nodes and hidden dimensions can be substantial.

\subsection{Real-Time Evaluation}
To evaluate  the  real-world applicability of domain adaptation methods, we conduct  a live deployment simulation by holding out the final  20\% of the target domain data (R-69) during training. This setup emulates  a streaming environment, where the model must make predictions on unseen, incoming data without access to  future information -- mirroring  operational requirements in air quality monitoring systems. As shown in Table~\ref{tab:rmse_results_O3_live}, domain adaptation techniques significantly reduce both normalized and actual error metrics compared  to the source-only baseline. TikUDA consistently achieves the lowest errors across all metrics, highlighting its strong generalization capabilities and robustness  to temporal distribution shifts. These results indicate that TikUDA is well-suited for deployment in dynamic, real-time environments, delivering  accurate and reliable  multisensor fusion without requiring full  retraining on the target domain.

\begin{table}[ht!]
  \centering
  \caption{Real-Time Evaluation Results for Domain Adaptation Scenario R-212 $\rightarrow$ R-69 (O$_3$)}
  \resizebox{\linewidth}{!}{
  \begin{tabular}{lcc|cc}
    \toprule
    \textbf{Method} & \textbf{RMSE (Norm.) \(\downarrow\)} & \textbf{RMSE (Actual) \(\downarrow\)} & \textbf{MAE (Norm.) \(\downarrow\)} & \textbf{MAE (Actual) \(\downarrow\)} \\
    \midrule 
    Source-Only & 0.315 & 73.57 & 0.284 & 66.22 \\
    MMD & 0.109 & 25.46 & 0.085 & 19.89 \\
    AdaBN & 0.262 & 61.20 & 0.214 & 50.03 \\
    CORAL & 0.177 & 41.20 & 0.153 & 35.61 \\
    ERM-NU & 0.132 & 30.71 & 0.108 & 25.13 \\
    AdaGCN & 0.098 & 22.96 & 0.075 & 17.52 \\
    DARE-GRAM & 0.088 & 20.63 & 0.070 & 16.23 \\
    TikUDA & \textbf{0.080} & \textbf{18.63} & \textbf{0.064} & \textbf{14.85} \\
    \bottomrule
  \end{tabular}
  }
  \label{tab:rmse_results_O3_live}
\end{table}

\subsection{Ablation Studies}
In this subsection, we explore the effects of aligning angles and scaling eigenvalues on model performance through ablation studies. We select the tasks R-212 $\rightarrow$ R-69 (O\textsubscript{3}) and R-69 $\rightarrow$ R-212 (O\textsubscript{3}) for this analysis. As illustrated in Table \ref{tab:ablation_studies}, aligning both angles and scales in the proposed method results in performance enhancements compared to the source-only approach. Notably, scaling alignment significantly reduces the RMSE compared to the source-only baseline by ensuring the range of predicted values aligns across different datasets. Meanwhile, aligning angles alone yields a smaller improvement in performance.

\begin{table}
  \centering
  \caption{RMSE (\(\downarrow\)) Values for Ablation Studies}
  \resizebox{0.75\linewidth}{!}{
    \begin{tabular}{lcc}
      \toprule
      Method & R-212 $\rightarrow$ R-69 (O\textsubscript{3}) & R-69 $\rightarrow$ R-212 (O\textsubscript{3}) \\
      \midrule
      Source-Only & 0.255 & 0.217 \\
      $\gamma_{\text{angle}} = 0$ & 0.096 & 0.106 \\
      $\gamma_{\text{scale}} = 0$ & 0.202 & 0.159 \\
      \textbf{Both (TikUDA)} & \textbf{0.087} & \textbf{0.099} \\
      \bottomrule
    \end{tabular}
  }
  \label{tab:ablation_studies}
\end{table}

To demonstrate the effectiveness of Haversine similarity in UDA tasks, we compare it against Cosine similarity based on their RMSE values in both adaptation directions between routes R-212 and R-69 (O\textsubscript{3}). As shown in Table~\ref{tab:havervscosine}, Haversine similarity achieves lower RMSE values, with 0.087 for adaptation from R-212 to R-69 and 0.097 for the reverse direction, indicating more consistent and accurate performance. In contrast, Cosine similarity results in higher RMSE values of 0.093 and 0.102, respectively. These results highlight that Haversine similarity provides better cross-domain alignment and is more effective for UDA.

\begin{table}
  \centering
  \caption{Comparison of RMSE (\(\downarrow\)) Values based on Similarity Functions}
  \resizebox{0.75\linewidth}{!}{
    \begin{tabular}{lcc}
      \toprule
      Similarity & R-212 $\rightarrow$ R-69 (O\textsubscript{3}) & R-69 $\rightarrow$ R-212 (O\textsubscript{3}) \\
      \midrule
      Haversine Similarity & \textbf{0.087} & \textbf{0.097} \\
      Cosine Similarity & 0.093 & 0.102 \\
      \bottomrule
    \end{tabular}
  }
  \label{tab:havervscosine}
\end{table}

\begin{figure}
\centering
\includegraphics[width=0.75\linewidth]{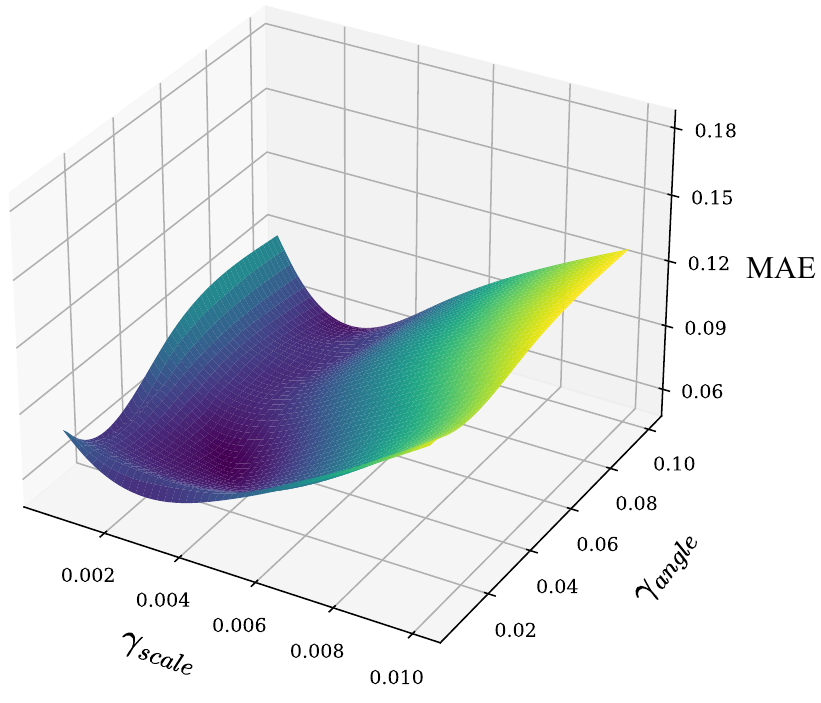}
\caption{Impact of $\gamma_{scale}$ and $\gamma_{angle}$ in MAE.}
\label{scaleangle}
\end{figure}

Figure \ref{scaleangle} illustrates the impact  of the hyperparameters $\gamma_{scale}$ and $\gamma_{angle}$ on model performance, measured  by MAE, for the R-212 $\rightarrow$ R-69 (O\textsubscript{3}) transfer case. The surface plot demonstrates that our method remains stable across a broad range hyperparameter combinations, with MAE values typically  ranging between 0.07 and 0.09. Notably, the model exhibits more sensitivity to variations  in $\gamma_{scale}$, likely because this parameter directly affects feature scaling, which plays a critical role in the adaptation process.  In contrast, $\gamma_{angle}$, which governs angular alignment, appears to   have a more limited effect  on performance  in this specific  task.

\begin{figure}
\centering
\includegraphics[width=\linewidth]{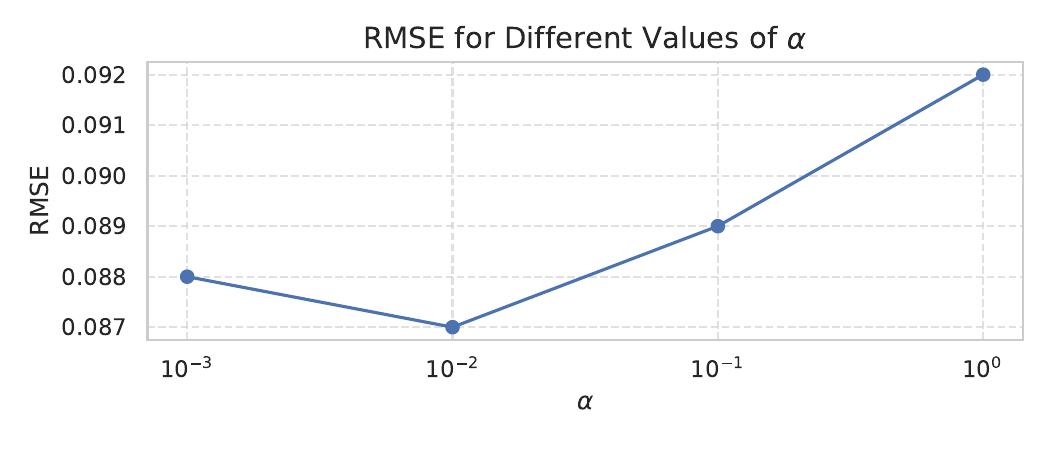}
\caption{Impact of Different Values of $\alpha$ on RMSE}
\label{alpha}
\end{figure}

The Tikhonov matrix introduces the hyperparameter $\alpha$, which plays a critical role in balancing numerical stability and effective alignment. When $\alpha$ is too low, the matrix may become  rank-deficient, potentially leading to numerical instability. Conversely, an excessively  large $\alpha$ can cause the identity matrix subspace to dominate  over  that of the Gram matrix, which undermines the alignment objective. As shown in Figure\ref{alpha}, our analysis of the R-212 $\rightarrow$ R-69 (O\textsubscript{3}) case shows that the performance of our method remains  largely stable across a wide range of $\alpha$ values. This indicates that the method is robust to the
 choice of $\alpha$, and that fine-tuning this parameter is not critical for achieving good results.

\section{Conclusion}
\label{sec:conclusion}
This paper addresses the challenge of unsupervised domain adaptation for spatial-temporal sensor fusion in spatial-temporal graph neural networks. Our proposed method, TikUDA, is inspired by the Tikhonov-regularized closed-form least squares solution to achieve effective domain adaptation for regression tasks. Additionally, we introduce an efficient scaling and alignment strategy for the Tikhonov matrix, which not only reduces computational complexity but also enhances model accuracy. Extensive experiments across different real-world scenarios and datasets, including air quality monitoring and EEG signal reconstruction, demonstrate that TikUDA significantly improves generalization across domains. These results validate the robustness, scalability, and versatility of our approach for spatial-temporal sensor networks. For future work, several promising research directions emerge from this work. An interesting direction for future work stems from the fact that TikUDA is inherently model-agnostic. While this paper focuses on STGNN-based architectures for spatial-temporal sensor data, the underlying domain adaptation framework is not limited to graph-based models. It would be valuable to explore how TikUDA performs with other backbone architectures commonly used in fields such as computer vision (e.g., CNNs, Vision Transformers) or multimodal learning frameworks. This could provide insights into its adaptability across different data modalities and model structures. Another potential direction involves investigating alternative regularization techniques beyond ridge regression that still yield closed-form solutions for inverse problems. Applying these advanced techniques to UDA tasks may further enhance performance and broaden applicability. Additionally, while our current framework assumes similar sensor configurations between domains, future work could incorporate adaptive techniques to enable flexible alignment between domains with differing graph sizes, thereby broadening TikUDA's applicability in more heterogeneous sensor deployment scenarios.

\appendix

\subsection{Details on Power Method}

The power iteration method is a widely used algorithm for finding the largest eigenvalue and its corresponding eigenvector of a matrix. It is particularly efficient for large matrices where computing the full spectrum of eigenvalues and eigenvectors (such as methods like SVD) would be computationally expensive. Power iteration approximates only the largest eigenvalue, making it suitable for tasks requiring less detailed spectral information. The algorithm begins by randomly initializing a vector \( \mathbf{v_0} \in \mathbb{R}^p \). In each iteration, the vector is updated by multiplying it by the matrix \( \mathbf{A} \), normalizing it to prevent it from growing unbounded. Specifically, for each iteration \( k \), the new vector \( \mathbf{v_k} \) is computed as: \begin{equation}
 \mathbf{v_k} = \frac{\mathbf{A v_{k-1}}}{\|\mathbf{A v_{k-1}}\|}, 
\end{equation} 
where \( \| \cdot \| \) represents the norm. Over several iterations, the vector \( \mathbf{v_k} \) converges to the eigenvector associated with the largest eigenvalue \( \lambda_{\text{max}} \). Once the algorithm converges, the largest eigenvalue can be approximated by:
\begin{equation}
 \lambda_{\text{max}} \approx \frac{\mathbf{v_k}^\mathsf{T} \mathbf{A v_k}}{\mathbf{v_k^\mathsf{T} v_k}}.  
\end{equation}

\section*{Acknowledgement}
This research was supported by the Swiss Federal Institute of
Metrology (METAS).

\bibliographystyle{IEEEtran}
\bibliography{IEEEabrv,main}





\end{document}